\documentclass[10pt,twocolumn,letterpaper]{article}

\usepackage[pagenumbers]{iccv} 

\usepackage{amsmath,amsfonts,bm}

\def\1{\bm{1}}

\DeclareMathAlphabet{\mathsfit}{\encodingdefault}{\sfdefault}{m}{sl}
\SetMathAlphabet{\mathsfit}{bold}{\encodingdefault}{\sfdefault}{bx}{n}

\newcommand{\R}{\mathbb{R}}

\newcommand{\ablation}[1]{#1}

\definecolor{mygreen}{rgb}{0.498,0.843,0.655}
\definecolor{myred}{rgb}{1.0,0.498,0.498}
\usepackage{tabularx}
\usepackage{multirow}
\usepackage{subcaption}
\usepackage{overpic}
\usepackage{stfloats}
\newcommand{\tabularxmulticolumncentered}[3] 
    {\multicolumn{#1}
                 {>{\centering\hsize=\dimexpr#1\hsize+#1\tabcolsep+\arrayrulewidth\relax}#2}
                 {#3}}
\usepackage{amsmath}
\DeclareFontFamily{U}{mathx}{}
\DeclareFontShape{U}{mathx}{m}{n}{<-> mathx10}{}
\DeclareSymbolFont{mathx}{U}{mathx}{m}{n}
\DeclareMathAccent{\widecheck}{0}{mathx}{"71}

\usepackage{amssymb}
\usepackage{pifont}
\usepackage{xcolor}
 \usepackage{tikz}
 
\usepackage{colortbl}
\newcommand{\inci}[1]{\includegraphics[width=1\linewidth]{#1}}

\newcommand{\incimi}[1]{\includegraphics[width=1\linewidth, trim= 2.12cm 1.4cm 2cm 2.68cm, clip]{#1}}

\newcommand{\incimicolor}[1]{
\begin{overpic}[angle=90,trim= 0cm 0.9cm 0cm 0.5cm,clip,height=1\linewidth]{#1}
    \put(5,3){\tiny 0}
    \put(5,97){\tiny $\nicefrac{1}{N}$}
\end{overpic}
}

\newcommand{\incimilabel}[1]{\includegraphics[height=1\linewidth, trim= 0cm 1.4cm 10.7cm 2.68cm, clip]{#1}}

\usepackage{nicefrac}       

\newcommand{\methodname}{GECO\xspace}

\usepackage[export]{adjustbox}
\definecolor{iccvblue}{rgb}{0.21,0.49,0.74}
\usepackage[pagebackref,breaklinks,colorlinks,allcolors=iccvblue]{hyperref}

\def\paperID{3956} 
\def\confName{ICCV}
\def\confYear{2025}

\title{GECO: Geometrically Consistent Embedding with Lightspeed Inference}

\author{
{Regine Hartwig \hspace{1cm}
Dominik Muhle\hspace{1cm}
Riccardo Marin\hspace{1cm}
Daniel Cremers}\\
TU Munich \hspace{1cm} Munich Center for Machine Learning\\
{\tt\small \{regine.hartwig, dominik.muhle, riccardo.marin, cremers\}@tum.de}
}

\begin{document}

\def\Ours{cnb1ilp9__} 
\def\Geo{geosc} 
\def\DIFT{dift_adm}
\def\DINOVtwo{dinov2_518_upft1}
\def\DINOVone{dino_224_upft4}
\def\GeoSC{geosc}
\newcolumntype{Y}{>{\centering\arraybackslash}X}
\newcolumntype{C}[1]{>{\centering\arraybackslash}p{#1}}

\twocolumn[{%
\renewcommand\twocolumn[1][]{#1}%
\maketitle
\begin{center}
    \centering
    \captionsetup{type=figure}
        \begin{overpic}[trim=0cm 0.0cm 0cm 0.0cm,clip, width=\linewidth]{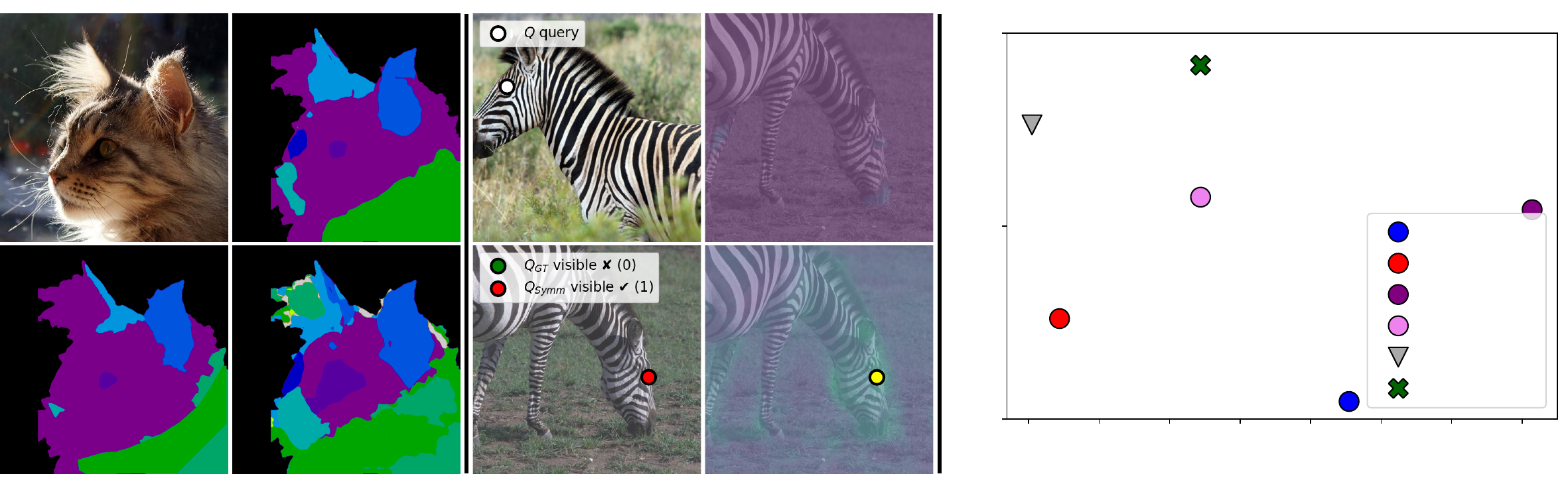}
	    \put(3, 30.5){\scriptsize Source Image}
            \put(6.5, -0.5){\scriptsize  GT}

	    \put(19, 30.5){\scriptsize  \methodname \textbf{(Ours)}}
            \put(20, -0.5){\scriptsize  Geo\cite{zhang2024telling}}

	    \put(33, 30.5){\scriptsize  Source Image}
            \put(34, -0.5){\scriptsize  Target Image}

	    \put(48.5, 30.5){\scriptsize  \methodname \textbf{(Ours)}}
            \put(50, -0.5){\scriptsize  Geo\cite{zhang2024telling}}

            \put(10, 31.8){\textbf{Segmentation}}
            \put(57, 31.8){\textbf{Keypoints Matching}}

            \put(90.3, 15.8){\scriptsize DINO}
            \put(90.3, 13.8){\scriptsize DIFT$_{\textit{adm}}$}
            \put(90.3, 11.8){\scriptsize DINOv2-S}
            \put(90.3, 9.8){\scriptsize DINOv2-B}
            \put(90.3, 7.8){\scriptsize Geo}
            \put(90.3, 5.8){\scriptsize \methodname \textbf{(Ours)}}

            \put(62.2, 3.5){\scriptsize 50}
            \put(62.2, 16) {\scriptsize 70}
            \put(62.2, 28.3) {\scriptsize 90}

            \put(61, 15.1){\rotatebox{90}{\scriptsize PCK}}
            \put(81, 1){\scriptsize fps}
            
            \put(65.3, 2.6){\scriptsize 0}
            \put(69.5, 2.6){\scriptsize 10}
            \put(74, 2.6){\scriptsize 20}
            \put(78.5, 2.6){\scriptsize 30}
            \put(83, 2.6){\scriptsize 40}
            \put(87.4, 2.6){\scriptsize 50}
            \put(92, 2.6){\scriptsize 60}
            \put(96.4, 2.6){\scriptsize 70}
            
            \end{overpic}
    \caption{We propose \textbf{\methodname}, an optimal-transport-based approach for learning geometrically consistent visual features. Characterizing geometric properties, like distinguishing left/right eyes or front/back legs, remains challenging even for sophisticated methods. For instance, in the keypoint transfer example, showing a zebra, Geo~\cite{zhang2024telling} confuses the eyes, while our features have low similarity.
    Our method learns robust feature representations, achieving high accuracy while remaining lightweight and efficient at inference time (plot on the right), enabling real-time applications at 30fps. Our feature embedding exhibits a continuous and structured understanding of objects, making it highly effective for image segmentation (left) and precise keypoint matching, even in challenging scenarios involving occlusions (middle, with attention maps overlaid on the target images).}

    \label{fig:teaser}
\end{center}
}]

\begin{abstract}
    Recent advances in feature learning have shown that self-supervised vision foundation models can capture semantic correspondences but often lack awareness of underlying 3D geometry. \methodname addresses this gap by producing geometrically coherent features that semantically distinguish parts based on geometry (e.g., left/right eyes, front/back legs). We propose a training framework based on optimal transport, enabling supervision beyond keypoints, even under occlusions and disocclusions. With a lightweight architecture, \methodname runs at 30 fps, 98.2\% faster than prior methods, while achieving state-of-the-art performance on PFPascal, APK, and CUB, improving PCK by 6.0\%, 6.2\%, and 4.1\%, respectively. Finally, we show that PCK alone is insufficient to capture geometric quality and introduce new metrics and insights for more geometry-aware feature learning.
    \href{https://reginehartwig.github.io/publications/geco/}{https://reginehartwig.github.io/publications/geco/}
\end{abstract}    
\section{Introduction}
\label{sec:intro}

Vision Foundation models either trained only on images~\cite{oquab2023dinov2, dhariwal2021diffusion} or text-image pairs~\cite{rombach2022high} produce flexible features that can be applied across various tasks such as image generation, style transfer, and correspondence estimation. While these models perform well across a range of applications, they are constrained by their limited ability to distinguish between geometric properties~\cite{el2024probing}. For instance, they may struggle to differentiate between left and right eyes or the legs of a chair, issues that have been explored in the context of the Janus problem~\cite{zhang2024telling, el2024probing}.
Since they are trained using consistency between self-augmented images, including flipping, they often do not differentiate between symmetric parts, and are even encouraged  to learn features that are invariant to such transformations.
These models often encounter challenges like ambiguous features for semantically similar regions and confusion due to occlusions. Such mismatches can lead to severe artifacts in practical downstream applications, such as a category template reconstruction pipeline incorrectly reconstructing 5-legged animals or chairs~\cite{monnier2022share}. More advanced approaches, such as Geo~\cite{zhang2024telling}, enhance SD+DINO features~\cite{zhang2024tale} to address left-right ambiguities. However, despite its advancements, this method is often too slow, taking several seconds for a single prediction, making it impractical and difficult to scale.

Our work proposes \textbf{GE}ometrically \textbf{CO}nsistent embeddings, an efficient and robust approach to address this problem without compromising efficiency. \ablation{A key limitation of previous methods is their reliance on argmax-based assignments during training, which inherently assumes the keypoint to be visible in both views and fails to account for \mbox{(dis-)occlusions}.
Our intuition is that it therefore does not provide a strong supervision signal for the model to learn meaningful features. Specifically, it overlooks the fact that we are dealing with a partial assignment problem. For instance, in cases of occlusion, annotated information is ignored during training, even though it could help the model learn the correct bin assignment. Furthermore, the supervision signal is sparse: only a few annotated points are used, while the rest of the image contributes nothing to the training process. Although soft-argmax assignments with Gaussian perturbations of the position have been proposed to mitigate this issue, they tend to introduce blurring and ultimately still depend on an argmax operation, which is not a natural fit for the underlying assignment problem.
Instead, to account for the partial assignments and differentiability, we introduce a novel \textit{soft assignment loss} that leverages optimal transport on top of vision foundation models, leading to a nuanced and discriminative feature learning process. This loss function provides strong supervision feedback, allowing our module to learn distinctive features that effectively differentiate symmetric keypoints.} The optimal transport loss is based on the Sinkhorn algorithm~\cite{cuturi2013sinkhorn}, which enables a differentiable soft assignment formulation for backpropagation.

Importantly, unlike previous approaches~\cite{sarlin2020superglue, sun2021loftr}, our method does not rely on cross-attention between image pairs, the marginal distributions of the optimal transport module are created and enforced differently (see supplementary), and the module operates without trainable parameters. As a result, our model is a \textit{feature encoder rather than a feature matcher}. \methodname focuses on robust representation learning rather than direct correspondence estimation between two images.

Concretely, we employ a pre-trained model and refine it through LoRA adaptation~\cite{hu2022lora}. 
Leveraging \textit{DINOv2} ensures computational efficiency, enabling more extensive data augmentation compared to prior methods and further enhancing the robustness of learned features.

Our method outperforms the state of the art while being \textbf{faster and smaller by two orders of magnitude w.r.t. the closest competitor} (40 ms vs 2127ms; 332MB vs 9GB). We perform an extensive evaluation, demonstrating our superiority on the classical PCK@0.1 metric on CUB, APK, and PFPascal datasets~\cite{ham2017proposal,zhang2024telling,wah2011caltech}. We also extend our analysis to complement the information provided by the PCK metric, which we found might not be indicative of some prediction modes. Thanks to this, we demonstrate that our better performance directly derives from a better geometrical understanding, which also leads to a more focused and calibrated prediction, distributing similarity on the correct area and assigning occluded parts to the bin (see \cref{fig:teaser}).

In summary, our contributions are:
\begin{enumerate}
    \item We propose a novel loss function and a lightweight architecture for image representation learning, leveraging optimal transport-based soft assignment;
    \item Our formulation leads to geometrically-aware features, enabling state-of-the-art performance on correspondence estimation while being significantly more efficient. It surpasses geometrically-aware competitors on multiple datasets while \emph{reducing computation time by 98\%}, maintaining the speed of the lightweight backbone.
    \item We conduct an extensive analysis of the common PCK metric and complement it with object part segmentation evaluation on PascalParts. Our method effectively separates parts, indicating that it learns meaningful, dense feature representations.
\end{enumerate}
\noindent
Our implementation will be instrumental for researchers interested in downstream tasks of geometry-aware encoders.

\section{Related Work}
\label{sec:related}

\paragraph{Deep features} 
Self-supervised and unsupervised methods have gained popularity~\cite{he2020momentum, chen2020simple, grill2020bootstrap, he2022masked, caron2021emerging, oquab2023dinov2, caron2018deep, fuest2024diffusion}, but if trained on image-level objectives~\cite{chen2020simple, grill2020bootstrap, he2022masked, caron2021emerging, caron2018deep} often fail to capture fine spatial details~\cite{tumanyan2022splicing, amir2021deep}. Patch-based methods like DINOv2~\cite{oquab2023dinov2}, inspired by~\cite{zhou2021ibot}, learn effective feature representations for tasks like clustering and matching. 
Diffusion models~\cite{dhariwal2021diffusion, rombach2022high} have become powerful generative models, with (Clean-)DIFT~\cite{tang2023emergent, stracke2025cleandift} enabling dense feature extraction for vision tasks. While DINOv1, DINOv2~\cite{caron2021emerging,oquab2023dinov2}, and DIFT~\cite{tang2023emergent} work well for zero-shot tasks~\cite{tumanyan2022splicing, amir2021deep, gupta2023asic, hedlin2023unsupervised, goodwin2022zero}, including semantic correspondence finding, they often struggle with geometric awareness, especially left-right distinctions~\cite{el2024probing,chen2025feat2gs}, potentially due to image-flipping augmentations~\cite{oquab2023dinov2}. Stable Diffusion (SD) features~\cite{dhariwal2021diffusion} encode more geometry than DINOv1~\cite{caron2021emerging}, but still lack left-right distinction, exposing limitations in spatial understanding, which can cause problems in 3D reconstruction~\cite{liu2023zero, khalid2022clip, monnier2022share}.

\paragraph{Geometry-aware matching}
Several methods adress keypoint matching for rigid objects, such as SuperPoint~\cite{detone2018superpoint} and SuperGlue~\cite{sarlin2020superglue}, which use homography supervision. LOFTR~\cite{sun2021loftr} improves upon this with dense optimal transport matching. These approaches rely on (cross-)attention and learnable optimal transport layers for sparse labels. Recent methods like DUST3R, MAST3R, Fast3R~\cite{wang2024dust3rb, leroy2024groundinga, yang2025fast3r} assume rigid motion, limiting their applicability to more complex cases.
Our method, \methodname, focuses on learning features rather than matching and handles deformable objects and complex geometries.
Deformable objects complicate training, as constraints (\eg epipolar) are not available. Improving matching of deformable objects, CATs++~\cite{cho2022cats++} uses attention layers, LCorrSan~\cite{huang2022learning} estimates dense flow via correlation layers,~\cite{cheng2024zeroshot} uses functional maps, and~\cite{truong2022probabilistic, li2021probabilistic} optimize dense probabilities. These methods focus on correspondence estimation. Our approach prioritizes feature learning, not matching.

\paragraph{Geometry-aware representation}
Recent work~\cite{yue2024improving, xu20243difftection, xiangli2025doppelgangers, wang2025vggtb} uses rigid inter-instance supervision for learning features with geometric awareness.
Others focus on geometry-aware deformable intra-instance representation learning:
Using 3D supervision, concurrent work~\cite{qian2025bridging,fundel2025distillation,kloepfer2024loco} achieve good results on the geometric-aware matching task and a similar speedup as our method. Unlike our focus on speed and efficiency,~\cite{xue2025matcha} targets matching performance using additional correspondence supervision and Diffusion Features. However, the need for extra signals and lack of real-time capability limit its practical applicability.
Earlier works like AnchorNet~\cite{novotny2017anchornet} use category-level supervision for geometry-aware features. DHF~\cite{luo2024diffusion} fine-tunes diffusion features for semantic matching with a sparse contrastive loss on keypoints. Unsup~\cite{thewlis2017unsupervised} and Sphere~\cite{mariotti2024improving} map to spherical coordinates with category and viewpoint supervision, while~\cite{dunkel2025yourself} also uses the spherical geometric prior without supervision. Completely unsupervised, SCOPS~\cite{hung2019scops} enforces equivariance for co-part segmentation, while~\cite{choudhury2021unsupervised} uses contrastive learning for part discovery.
We consider Geo~\cite{zhang2024telling} the most closely related work to ours. It uses a sparse contrastive loss with keypoint annotations and a soft argmax operator for feature enhancement. However, its reliance on SD features results in slower inference, category annotation dependency, and larger model size. Additionally, the extra parameters introduced by their head are significant, increasing the potential for overfitting.

\paragraph{Training with sparse correspondences}
To leverage sparse correspondence during training, a common approach applies an argmax operator~\cite{zhang2024telling,luo2024diffusion} with a sparse contrastive loss~\cite{ilharco2021openclip, radford2021learning}. While effective in some domains, this method, when used with keypoint annotations~\cite{luo2024diffusion}, provides a learning signal to only a small subset of annotated patches visible in both images, resulting in suboptimal feature representations~\cite{zhang2024telling}.
We leverage optimal transport (OT)~\cite{cuturi2022tutorial, cuturi2013sinkhorn} not as a matching module, but to define a structured soft assignment loss. Specifically, we employ the Sinkhorn algorithm~\cite{cuturi2022tutorial, cuturi2013sinkhorn, pele2009fast}, a differentiable OT solver that produces dense supervision across all image patches~\cite{bonneel2023survey, izquierdo2024optimal, xing2022differentiable} by minimizing the cost to transport mass between distributions. Unlike prior work~\cite{sarlin2020superglue, sun2021loftr}, our formulation requires no additional trainable parameters, resulting in more broadly applicablel and task-agnostic features.

\paragraph{Our positioning} Our method builds on a lightweight and efficient DINOv2-B backbone, fine-tuned using LoRA~\cite{hu2022lora} to improve both computational efficiency and memory usage. To address the core challenges of sparse annotations and \mbox{(dis-)occlusions} in loss design, we introduce a soft optimal transport layer. This layer delivers a learning signal to all patches, effectively utilizing the full set of available annotations and enabling the learning of robust and discriminative feature representations.
\section{Background}
\label{sec:background}

\paragraph{Argmax matching}\label{sec:background:argmax_matching}
Considering a source image $S$ and target one $T$, both equipped with a set of patches represented by
their indices $\mathcal{I}:=\{i_1,...,i_l\}$ and $\mathcal{J}:=\{j_1,...,j_m\}$ and
their d-dimensional features $\mathbf{X}^s\in\R^{l\times d}$ and $\mathbf{X}^t \in \R^{m\times d}$ respectively. Given a query location $i$, the corresponding $\widehat{j}$ on the target can be received by:
\begin{equation}\label{eq:matching}
    \widehat{j} = \arg \max_j \left< \mathbf{X}^s_i, \mathbf{X}^t_j\right>.
\end{equation}

While simple, argmax matching suffers from several drawbacks. It introduces ambiguity, as multiple geometrically distinct source locations can map to the same target point (e.g., left and right eyes both matching to one eye). It does not account for occlusions, often resulting in incorrect assignments with low similarity. Moreover, it imposes hard, one-hot assignments that assume equal mass per patch across images, limiting its ability to handle scale differences or partial matches. Crucially, it ignores the global structure of the assignment, focusing only on local maxima, which leads to sparse gradients and hinders effective training. We instead formulate matching as an optimal transport problem. To handle occlusions, we introduce a dustbin entry in the assignment matrix, and employ a soft, iterative solver~\cite{cuturi2013sinkhorn,cuturi2022tutorial} that provides meaningful gradients for all features, improving training stability and convergence.

\paragraph{Optimal Transport formulation}
Suppose we have pairs of patches
$(i, j) \in \mathcal{I} \times \mathcal{J}$.
We solve for an assignment between two images~\cite{oquab2023dinov2, sarlin2020superglue} by using the cosine similarity of the features as score matrix $\mathbf{C}$ with 
\begin{equation}
\mathbf{C}_{i,j} = \left<\mathbf{X}^s_i,\mathbf{X}^t_j\right> \in [-1,1].
\end{equation}
In order to model occlusions properly, we augment the vectors with a dustbin at $l'=l+1$ and $m'=m+1$, which gets assigned a threshold parameter $z\in \R $:
\begin{equation}
 \mathbf{C}_{i_{l'},i_{m'}} = \mathbf{C}_{i_{l'},j}= \mathbf{C}_{i,i_{m'}} = z \quad \forall i,j \in \mathcal{I} \times \mathcal{J}.
\end{equation}
We then solve for an assignment matrix $\mathbf{P} \in U(\mathbf{a},\mathbf{b})$, where
\begin{equation}
U(\mathbf{a},\mathbf{b}) := \left\{ \mathbf{P} \in \R_+^{i_{l'}\times i_{m'}}| \mathbf{P} \mathbf{1}_{i_{m'}} = \mathbf{a}, \mathbf{P}^T \mathbf{1}_{i_{l'}} = \mathbf{b} \right\},
\end{equation}
$\mathbf{a} \in \Sigma_{i_{l'}}$, and $\mathbf{b}\in \Sigma_{i_{m'}}$, i.e. the marginalizations of $P$ (see supplementary) are in the respective simplex \smash{$\Sigma_n := \{\mathbf{x}\in \R_+^n: \mathbf{x}^T \mathbf{1}_n=1\}$}. The assignment can be obtained by solving the OT problem
\begin{equation}\label{eq:matching_linear}
    \widehat{\mathbf{P}} = \arg\max_{\mathbf{P} \in U(\mathbf{a},\mathbf{b})} \left< \mathbf{P}, \mathbf{C} \right>.
\end{equation}
To provide gradients for all input features during training, we use a regularized OT, which yields a smooth and differentiable assignment
\begin{equation}\label{eq:matching_ot_soft}
    \widehat{\mathbf{P}}^{\lambda} = \arg\max_{\mathbf{P}\in U(\mathbf{a},\mathbf{b})} \left< \mathbf{P}, \mathbf{C} \right> + \lambda H(\mathbf{P}),
\end{equation}
with $H(\mathbf{P}) = -\sum_{i,j} \mathbf{P}_{ij} \log \mathbf{P}_{ij}$ being the entropy of the assignment matrix. For $\lambda>0$ 
the entropy regularization promotes smoother, less sparse assignments, resulting in a differentiable soft assignment $\widehat{\mathbf{P}}^{\lambda}$ that provides a gradient signal for all input features.
Furthermore, changing the problem to an unbalanced OT problem, using additionally the KL-divergence as a regularizer~\cite{peyre2019computational}, helps in the case of unbalanced marginals $\mathbf{a}$ and $\mathbf{b}$,
meaning, that the amount of mass that needs to be assigned is not equal for both marginals
\begin{align}\label{eq:matching_ot_soft_kl}
 \widehat{\mathbf{P}}^{\lambda,\alpha, \beta} = &\arg\max_{\mathbf{P}\in \R_+^{i_{l'} \times i_{m'}}} \left< \mathbf{P}, \mathbf{C} \right> + \lambda H(\mathbf{P}) \nonumber \\
 &- \alpha \operatorname{KL}\left(\mathbf{P} \mathbf{1}_{i_{m'}} || \mathbf{a}\right) - \beta \operatorname{KL}\left(\mathbf{P}^T \mathbf{1}_{i_{l'}} || \mathbf{b}\right).
\end{align}
\ablation{In general, for formulating the OT problem, we need to know the distributions of $\mathbf{a}$ and $\mathbf{b}$.
However, using the unbalanced OT problem, we can relax the requirement of knowing the marginals, which is crucial for our application, as we do not have access to the ground truth marginals, but only estimates.}
\section{Method}
\label{sec:method}
\begin{figure*}[t]
    \centering
    \vspace{0.5cm}
    \footnotesize
    \begin{overpic}[trim=0cm 0.0cm 0cm 0.0cm,clip,width=\linewidth]{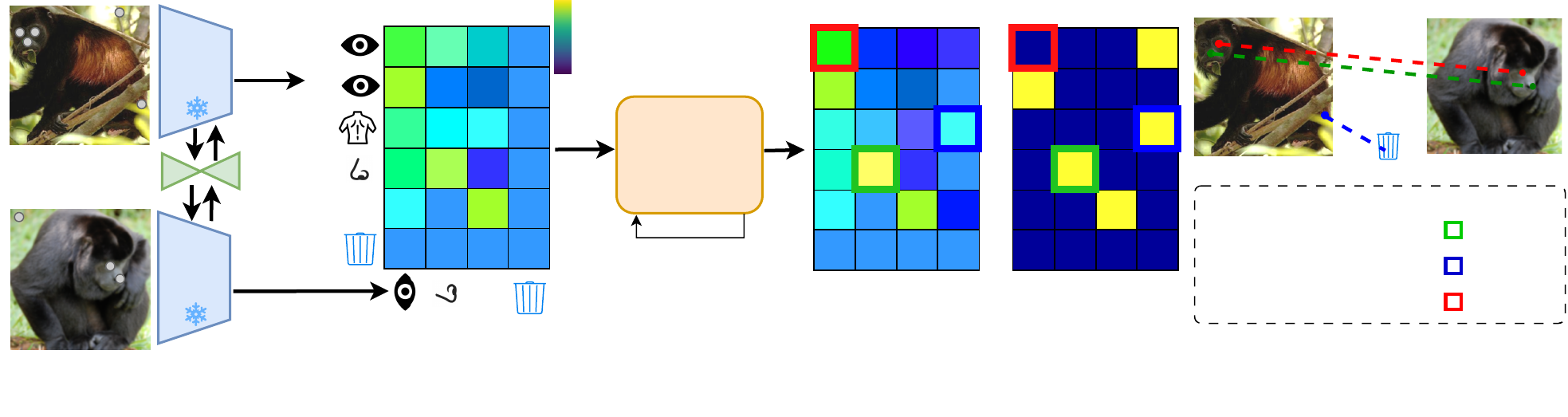}
        \put(37,25.8){$=$}
        \put(37,22){$\neq$}
        \put(5,27.8){\makebox(0,0){$S$}}
        \put(5,15){\makebox(0,0){$T$}}
        \put(44,18){\makebox(0,0){Differentiable}}
        \put(44,16){\makebox(0,0){Sinkhorn}}
        \put(44,14){\makebox(0,0){Algorithm}}
        \put(43,10){$10 \times$}
        \put(21.5,13){Bkg}
        \put(30.5,6){\rotatebox{90}{Bkg}}
        \put(15.5,15){LoRA}
        \put(12.5,23){\makebox(0,0){DINO}}
        \put(12.5,21.5){\makebox(0,0){v2}}
        \put(12.5,10){\makebox(0,0){DINO}}
        \put(12.5,8.5){\makebox(0,0){v2}}
        \put(29,26.8){\makebox(0,0){Cosine Similarity}}
        \put(57.5,27.8){\makebox(0,0){Optimized}}
        \put(57.5,25.8){\makebox(0,0){Assignment}}
        \put(70,27.8){\makebox(0,0){GT}}
        \put(70,25.8){\makebox(0,0){Assignment}}
        \put(94,12){\tiny positive pair}
        \put(94,10){\tiny positive pair}
        \put(94,9){\tiny with bin}
        \put(94,7){\tiny negative pair}
        \put(76.5,12){\scalebox{0.6}{$\mathcal{L} = $}}
        \put(78.5,12){\scalebox{0.6}{$- \sum_{(i,j)\in \mathcal{M}^+} \log  \widehat{\mathbf{P}}^{\lambda,\alpha, \beta}_{i,j}$}}
        \put(78.5,9.5){\scalebox{0.6}{ $- \sum_{(i,j)\in \mathcal{M}^0} \log  \widehat{\mathbf{P}}^{\lambda,\alpha, \beta}_{i,j}$}}
        \put(78.5,7){\scalebox{0.6}{ $-\sum_{(i,j)\in \mathcal{M}^-} \log (1- \widehat{\mathbf{P}}^{\lambda,\alpha, \beta}_{i,j})$}}
    \end{overpic}
    \vspace{-1cm}
    \caption{\textbf{Architecture and Training.} At training time, our method takes a pair of images composed of a source $S$ and targets one $T$, and obtains features through a DINOv2~\cite{oquab2023dinov2} frozen model complemented with a LoRA adapter~\cite{hu2022lora}.
    The features are used to compute the matching by cosine similarity associations between all patches. For visualization purposes, we show the matrix for the landmarks (grey dots). We pass the obtained matrix to a differentiable Optimal Transport Layer that, in a few iterations, obtains the predicted assignment. We compare this with the ground truth, and we use this as a supervision signal for our method. Our training objective also provides supervision signals on negative keypoint pairs and, thanks to the ``bin'' category, on keypoints that do not even have a visible counterpart in one of the two images.}
    \label{fig:architecture}    
\end{figure*}

We now describe our approach, including the feature learning module and the differentiable OT layer without learnable parameters, which processes the cosine similarity of patches. Finally, we introduce a loss function that encourages higher output probabilities for positive and bin pairs and lower probabilities for negative pairs.

\paragraph{Architecture}\label{sec:method:architecture}
Our architecture is simple, efficient, and interpretable (see \cref{fig:architecture}). We use a pre-trained DINOv2-B model~\cite{oquab2023dinov2}, kept frozen during training, and adapt it with a LoRA adapter~\cite{hu2022lora} of rank 10.

\paragraph{Differentiable OT layer}\label{sec:method:otlayer}
We compute the cosine similarity of all features from $S$ and $T$, including the background, obtaining a cosine similarity matrix. We augment this matrix with an additional row and column, setting a bin threshold of $z = 0.3$, and obtaining the score Matrix $\mathbf{C}$.

Next, we pass $\mathbf{C}$ to a differentiable optimal transport (OT) layer, which computes the optimal transport plan between the features of $S$ and $T$. Unlike previous works~\cite{sarlin2020superglue, sun2021loftr}, which enforce hard constraints on the boundaries $\mathbf{a}$ and $\mathbf{b}$, we adopt a KL-regularized soft assignment (\cref{eq:matching_ot_soft_kl}), allowing greater flexibility. Additionally, the OT layer has no learnable parameters and contributes to the loss function rather than being part of the model itself.

The marginals $\mathbf{a}$ and $\mathbf{b}$ of the multivariate probability matrix $\widehat{\mathbf{P}}$ are estimated using mask annotation and a method to determine the amount of the visible mass of the shape. Details on this are provided in the supplementary material.
The Sinkhorn-Knopp algorithm runs for $10$ iterations, producing a probability matrix $\widehat{\mathbf{P}}^{\lambda, \alpha, \beta}$ with the same dimensions as the score matrix $\mathbf{C}$.
The hyperparameters of the OT layer are $\lambda = 0.1$, $\alpha = 10$, and $\beta = 10$.



\paragraph{Binary cross entropy loss}\label{sec:method:loss}
As we do not have access to the ground truth matrix $\mathbf{P}$, which assigns all features of $S$ to all features of $T$, but only to the sets of positive, negative, and bin correspondences $\mathcal{M}^+, \mathcal{M}^-, \mathcal{M}^0 \subset \mathcal{I}\times \mathcal{J}$, we formulate our loss only on these sets.
Using a binary cross-entropy loss, we train the model to predict the correct values at some sparse entries of the assignment matrix $\mathbf{P}$ by
\begin{equation}
 \mathcal{L} = - \sum_{(i,j)\in \mathcal{M}^+ \cup \mathcal{M}^0} \log  \widehat{\mathbf{P}}^{\lambda,\alpha, \beta}_{i,j} \, -\sum_{(i,j)\in \mathcal{M}^-} \log (1- \widehat{\mathbf{P}}^{\lambda,\alpha, \beta}_{i,j}).
\end{equation}
We additionally add matches between foreground and background features to the set $\mathcal{M}^-$, which are not part of the ground truth, to enforce the model to learn to distinguish between the two.

\section{Experiments}
In this section we extensively evaluate our method compared to baselines, analyzing their capability of understanding geometry. First, in \cref{sec:experiments:pck}, we utilize the established PCK metric on different datasets, where we show that we are on par and better than the previous state-of-the-art. We also provide a detailed analysis of the limitations of PCK, and propose new metrics to inspect geometric knowledge of the features. In \cref{sec:experiments:segmentation}, we show that our method, using centroid clustering, also performs well on the pixel-level classification task without additional training, relying on a simple supervised classifier. This indicates that our approach effectively learns meaningful features.
\subsection{Analysis with (and of) PCK}\label{sec:experiments:pck}

\begin{figure}[t]
  \centering
  
  \begin{overpic}[width=0.55\linewidth]{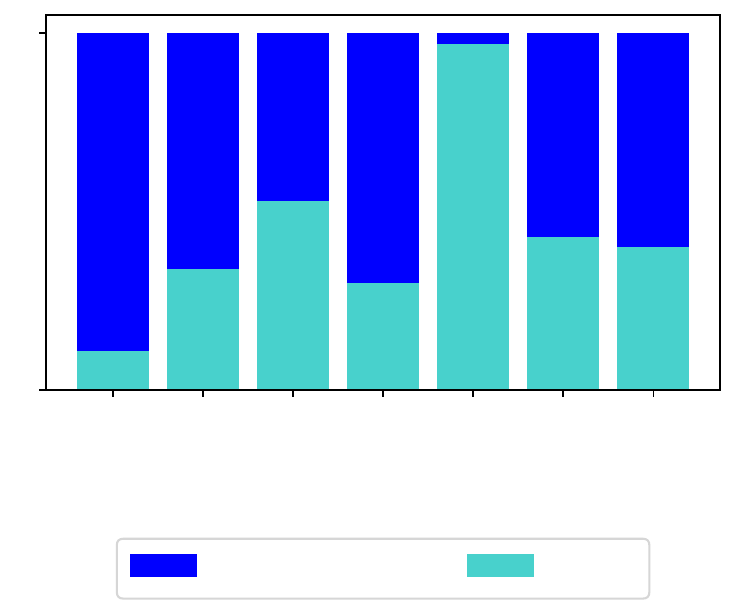}
    \put(1,78){\tiny{$1$}}
    \put(1,29){\tiny{$0$}}
    \put(15,28){\makebox(0,0){\tiny Bird}}
    \put(27.5,28){\makebox(0,0){\tiny Aero}}
    \put(40,28){\makebox(0,0){\tiny Bike}}
    \put(52.4,28){\makebox(0,0){\tiny Boat}}
    \put(64.9,28){\makebox(0,0){\tiny Bott}}
    \put(77.4,28){\makebox(0,0){\tiny Bus}}
    \put(89,28){\makebox(0,0){\tiny Car}}
    \put(29,5.2){\scalebox{0.6}{$\nicefrac{(n_{10}+n_{1x})}{n}$}}
    \put(76,5.2){\scalebox{0.6}{$\nicefrac{n_{11}}{n}$}}
    \put(11.2,23.5){\tikz \draw[black] (0,0)--(0.4,0);} 
    \put(23.3,23.5){\tikz \draw[black] (0,0)--(3.2,0);} 
    \put(12.1,20){\tiny{CUB}}
    \put(12.6,16){\tiny{\cite{wah2011caltech}}}
    \put(50,20){\tiny{Spair}}
    \put(50.4,16){\tiny{~\cite{min2019spair}}}
  \end{overpic}
  \begin{overpic}[width=0.4\linewidth]{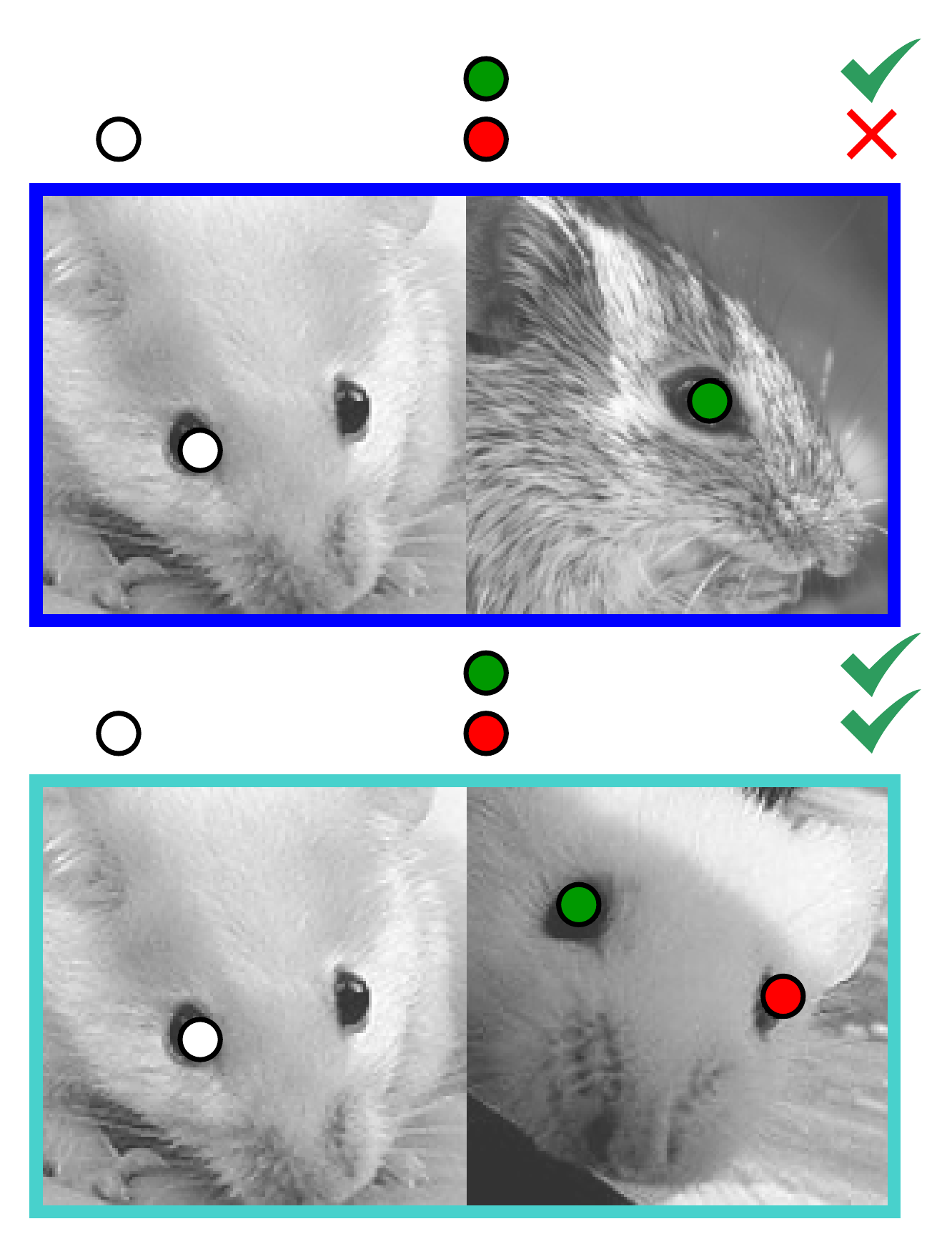}
    \put(12,87.8){\tiny{$Q$}}
    \put(17,87.8){\tiny{Query}}
    \put(41.4,92.5){\tiny{$Q_{\text{GT}}$}}
    \put(52.4,92.5){\tiny{visible(1)}}
    \put(41.4,87.8){\tiny{$Q_{\text{sym}}$}}
    \put(52.4,87.8){\tiny{visible(0)}}
    \put(12,40.3){\tiny{$Q$}}
    \put(17,40.3){\tiny{Query}}
    \put(41.4,45){\tiny{$Q_{\text{GT}}$}}
    \put(52.4,45){\tiny{visible(1)}}
    \put(41.4,40.3){\tiny{$Q_{\text{sym}}$}}
    \put(52.4,40.3){\tiny{visible(1)}}
  \end{overpic}
  \caption{\textbf{PGCK Dataset imbalance}. We report keypoint pairs grouping under our PGCK subdivision. $n_{11}$ counts only the pairs for which a geometric mismatch is possible. Other keypoint pairs (having no geom. counterpart/ with occluded geom. counterpart) are counted in $(n_{10}+n_{1x})$. Due to the high category imbalance, geometric error modes impact the overall PCK differently.}\label{fig:subsets_PCK}
\end{figure}

\paragraph{Percentage of correct keypoints (PCK)}
The percentage of correct keypoints (PCK) is a widely used metric for evaluating the performance of correspondence estimation methods.
We take an image pair, where keypoint matches are annotated, and evaluate how many query points $Q$ are correctly reprojected into the second view (see \cref{fig:pck_quanti1,fig:pck_quanti2}).
A reprojection is labeled as correct if the distance $\epsilon$ between predicted and annotated location $Q_{\text{GT}}$ is less than a threshold $\alpha_{img}$ of the image size or $\alpha_{bbox}$ of the bounding box size.
Following the evaluation protocol of previous work~\cite{tang2023emergent}, we use argmax matching for building assignments and reprojection only in one direction. 
In this work, we consider the PCK\texttt{point}@$\alpha_{img}$ version of it, where the percentage of correct keypoint reprojections per point is $\nicefrac{\widehat{n}}{n}$, the fraction of the sum of correctly reprojected points $\widehat{n}$ across all image pairs to all annotated point pairs $n$. 
It is interesting to note that PCK considers only keypoint pairs in which the query keypoint is visible in the target image. Hence, it does not evaluate a model's performance when the reprojection is not visible in the target image, but its symmetric counterpart is (see \cref{fig:pck_quanti1}, first four rows). To address this, we also report performance using qualitative results.

\paragraph{Percentage of geometry-aware correct keypoints (PGCK)}
Although PCK has played an important role in evaluating matching methods, we believe it is insufficient to depict the methods' understanding of geometry. We propose to break down the proportion between the correct reprojections $\widehat{n}$ and the total number of keypoints $n$ into different sets. Specifically, we separate the evaluation of query points that have a visible symmetric counterpart in the target image ($n_{11}$, an example in \cref{fig:subsets_PCK}, second row) from those that have a symmetric counterpart but are occluded in the target, and those for which a symmetric counterpart does not exist ($n_{10}$ and $n_{1x}$ respectively; example of the first in \cref{fig:subsets_PCK}, first row). PCK can then be divided in:
\begin{align}
    \text{PCK\texttt{point}} &= \frac{\widehat{n}}{n} \\
    &= \frac{\widehat{n}_{10} + 
    \widehat{n}_{11} + \widehat{n}_{1x}}{{n}_{10} + {n}_{11} + {n}_{1x}}\\
     &= 
     \underbrace{\frac{\widehat{n}_{10}}{n_{10}} \frac{n_{10}}{n} }_{\substack{Q_{\text{GT}}\text{\checkmark} \\ Q_{\text{Symm}}\text{\ding{55}} }
     } 
     + \underbrace{
      \frac{\widehat{n}_{11}}{n_{11}} \frac{n_{11}}{n}}_{\substack{Q_{\text{GT}}\text{\checkmark} \\ Q_{\text{Symm}}\text{\checkmark} \\[0.5em]\text{PGCK}}
      } 
     + \underbrace{
      \frac{\widehat{n}_{1x}}{n_{1x}} \frac{n_{1x}}{n}}_{\substack{Q_{\text{GT}}\text{\checkmark} \\ Q_{\text{Symm}}\text{-} }
      },
\end{align}
where the number of keypoint pairs $n = n_{10} + n_{11} + n_{1x}$  and number of correct reprojections $\widehat{n} = \widehat{n}_{10} + \widehat{n}_{11} + \widehat{n}_{1x}$.
Although PCK comprehends all these quantities, we see that evaluation on $n_{10}$ and $n_{1x}$ is less informative than those on set $n_{11}$, where actually the method could get confused by the presence of a symmetric element. For evaluating the geometric reasoning, we are interested only in the number $\widehat{n}_{11}$, where the model correctly matches the query keypoint when the symmetric counterpart is also visible in the target image. We can see its relevance by measuring the distribution of the annotated point pairs for the CUB and SPair datasets in the three categories. We report the counting in \cref{fig:subsets_PCK}. In the CUB dataset, $\nicefrac{n_{1x}}{n} =78\%$ of the keypoint pairs have no symmetric counterpart. For the remaining 22\% of the keypoint pairs, in only 11\% of the keypoint pairs, the symmetric counterpart is also visible in the second view. The SPair dataset has a high variation between the categories, with the highest value for the bottle category (97\%) and the lowest for the bird category (23\%).
In the following, we report both evaluations using PCK metric and our proposed split. We call such division \emph{Percentage of geometry-aware correct keypoints} (PGCK).

\paragraph{Geometric Ambiguity}

\begin{figure}
  
  \begin{overpic}[width=\linewidth]{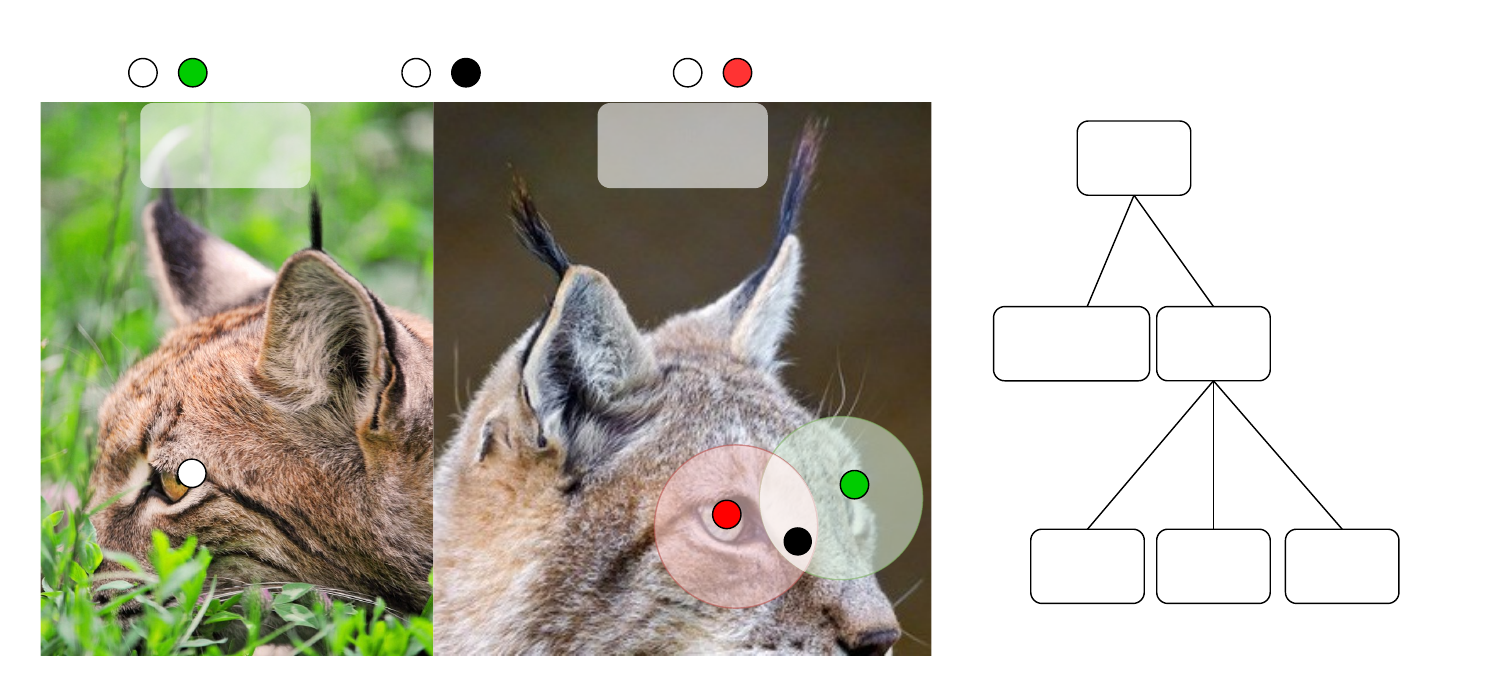}
    \put(7,39.2){\scalebox{0.6}{$(\quad,\quad)\in \underline{M}$}}
    \put(25.25,39.2){\scalebox{0.6}{$(\quad,\quad)\in \tilde{M}$}}
    \put(43.5,39.2){\scalebox{0.6}{$(\quad,\quad)\in \overline{M}$}}
    \put(11,34.5){\tiny{Source $S$}}
    \put(41.5,34.5){\tiny{Target $T$}}
    \put(75,34.5){\makebox(0,0){\scalebox{0.6}{$\widehat{n}$}}}
    \put(71.1,22){\makebox(0,0){\scalebox{0.6}{$\widehat{n}_{1x}+\widehat{n}_{10}$}}}
    \put(80.5,22){\makebox(0,0){\scalebox{0.6}{$\widehat{n}_{11}$}}}
    \put(72,7){\makebox(0,0){\scalebox{0.6}{$\underline{n}_{11}$}}}
    \put(80.5,7){\makebox(0,0){\scalebox{0.6}{$\tilde{n}_{11}$}}}
    \put(89.2,7){\makebox(0,0){\scalebox{0.6}{$\overline{n}_{11}$}}}
  \end{overpic}
  \caption{\textbf{Geometric Ambiguity of PGCK.} In the case of symmetric correspondences, the PCK metric does not account for ambiguous assignments of true positive correspondences. The {\color{mygreen} Green dot} around the annotated match will collect an assignment that results in a true positive match, the {\color{myred} Red dot} collects symmetric mismatches. In cases where the symmetric counterpart keypoint is sufficiently close, the circles overlap. These points can not be unambiguously assigned to either keypoint. The PCK metric still counts them as true positives. Points falling into the ambiguous set area are collected in $\tilde{M}$ and make up around $50 \%$ of the prediction for all investigated methods.
  }
  \label{fig:subsets_PGCK}
\end{figure}

Although $\nicefrac{\widehat{n}_{11}}{n_{11}}$ show informativeness about geometric knowledge of models, to complete our analysis, we also highlight a special set that is worth further investigation. Specifically, the set $\tilde{M}$ with $|\tilde{M}|=\tilde{n}_{11}$ contains point pairs where the predicted location is close (with a distance smaller than the defined radius) to $Q_{\text{GT}}$ and $Q_{\text{Symm}}$ as shown in \cref{fig:subsets_PGCK}.

\begin{table}[t]
  \footnotesize
    \centering 
    \begin{tabularx}{1\linewidth}{m{1.9cm} *{3}{Y}}
      & \cellcolor{mygreen!30}Unambiguous Correct Pred.
      & ambiguous & \cellcolor{myred!30}Unambiguous Wrong Pred.\\[1.3ex]
       
        & \centering \cellcolor{mygreen!30}$\frac{\underline{n}_{11}}{n_{11}}\uparrow$
        & \centering$\frac{\tilde{n}_{11}}{n_{11}}$
        & \cellcolor{myred!30}$\frac{\overline{n}_{11}}{n_{11}}\downarrow$\\[1.3ex]
        \hline
        DINO~\cite{caron2021emerging}              & 17.0 & 38.8 & 16.0\\
        DIFT$_{\text{ad}}$~\cite{tang2023emergent}  & 25.2 & 43.8  & 8.4\\
        DINOv2-S~\cite{oquab2023dinov2}              & 25.1 & 50.9 & 13.8\\
        DINOv2-B~\cite{oquab2023dinov2}              & 24.5 & 51.8 & 16.0\\
        \hline
        Geo~\cite{zhang2024telling} & 36.2 & 53.1 & 2.9\\
        \methodname (Ours)  & {\textbf{40.0}} &  {53.2}  & {\textbf{2.3}} \\ 
        \hline
    \end{tabularx}
    \caption{\textbf{Geometric ambiguity} Analysis of PGCK on APK~\cite{zhang2024telling}. We outperform previous work in the unambiguous geom. correct matching (left) by 3.8 \%, while our method disregards more of the unambiguously wrong pairs (right). The best scores are highlighted in \textbf{bold}.}\label{tab:pgck_splits}
\end{table}
Therefore, success in these cases does not directly measure geometric knowledge, as it ``cheats'' the metric by predicting points in the middle of the two. To compensate for this, it is possible to derive two further measures. First, the \textbf{unambiguous true positives U-TP} $\underline{n}_{11}/ n_{11}$, which measures the cardinality $|\underline{M}|=\underline{n}_{11}$ of the subset, which contains correctly matched keypoint pairs, where $Q_{\text{Symm}}$ is far enough away from the predicted position.
Second, the \textbf{false correspondences} $\overline{n}_{11}/ n_{11}$, which measures the cardinality $|\overline{M}|=\overline{n}_{11}$ of the subset, which contains wrongly matched keypoint pairs, where $Q_{\text{GT}}$ is far enough away from $Q_{\text{Symm}}$ to be considered as a completely wrong match.
As we will see later, the high number of keypoint pairs $\tilde{n}_{11}$ indicates that the commonly used radius $\alpha_{img}=0.1$ is too big for the PCK metric. A further analysis of how the subsets behave for other values of $\alpha_{img}$ is provided in the supplementary material.

\setlength{\tabcolsep}{2pt}  
\renewcommand{\arraystretch}{1}  
\newcommand{\QualiRowA}[3]{
    \inci{figures/05_experiments/pck/figures_pck/#1/dinov2_b14_518_upft1/src_01_#2_crop.jpg}
    &\inci{figures/05_experiments/pck/figures_pck/#1/dinov2_b14_518_upft1/trg_01_#2_gt_crop.jpg}
    &\inci{figures/05_experiments/pck/figures_pck/#1/dinov2_b14_518_upft1/trg_01_#2_crop.jpg}
    &\inci{figures/05_experiments/pck/figures_pck/#1/\Geo/trg_01_#2_crop.jpg}
    &\inci{figures/05_experiments/pck/figures_pck/#1/\Ours/trg_01_#2_crop.jpg}\\
}
\newcommand{\QualiRowACUB}[3]{
    \inci{figures/05_experiments/pck/figures_pck/#1/dinov2_b14_518_upft1/src_01_#2_crop.jpg}
    &\inci{figures/05_experiments/pck/figures_pck/#1/dinov2_b14_518_upft1/trg_01_#2_gt_crop.jpg}
    &\inci{figures/05_experiments/pck/figures_pck/#1/dinov2_b14_518_upft1/trg_01_#2_crop.jpg}
    &\inci{figures/05_experiments/pck/figures_pck/#1/\Geo/trg_01_#2_crop.jpg}
    &\inci{figures/05_experiments/pck/figures_pck/#1/\Ours/trg_01_#2_crop.jpg}\\
}

\newcommand{\QualiRowB}[3]{
    \inci{figures/05_experiments/pck/figures_pck/#1/dinov2_b14_518_upft1/src_11_#2_crop.jpg}
    &\inci{figures/05_experiments/pck/figures_pck/#1/dinov2_b14_518_upft1/trg_11_#2_gt_crop.jpg}
    &\inci{figures/05_experiments/pck/figures_pck/#1/dinov2_b14_518_upft1/trg_11_#2_crop.jpg}
    &\inci{figures/05_experiments/pck/figures_pck/#1/\Geo/trg_11_#2_crop.jpg}
    &\inci{figures/05_experiments/pck/figures_pck/#1/\Ours/trg_11_#2_crop.jpg}\\
}
\newcommand{\QualiRowBCUB}[3]{
    \inci{figures/05_experiments/pck/figures_pck/#1/dinov2_b14_518_upft1/src_11_#2_crop.jpg}
    &\inci{figures/05_experiments/pck/figures_pck/#1/dinov2_b14_518_upft1/trg_11_#2_gt_crop.jpg}
    &\inci{figures/05_experiments/pck/figures_pck/#1/dinov2_b14_518_upft1/trg_11_#2_crop.jpg}
    &\inci{figures/05_experiments/pck/figures_pck/#1/\Geo/trg_11_#2_crop.jpg}
    &\inci{figures/05_experiments/pck/figures_pck/#1/\Ours/trg_11_#2_crop.jpg}\\
}
\newcommand{\QualiRowBPF}[3]{
    \inci{figures/05_experiments/pck/figures_pck/#1/dinov2_b14_518_upft1/src_1x_#2_crop.jpg}
    &\inci{figures/05_experiments/pck/figures_pck/#1/dinov2_b14_518_upft1/trg_1x_#2_gt_crop.jpg}
    &\inci{figures/05_experiments/pck/figures_pck/#1/dinov2_b14_518_upft1/trg_1x_#2_crop.jpg}
    &\inci{figures/05_experiments/pck/figures_pck/#1/\Geo/trg_1x_#2_crop.jpg}
    &\inci{figures/05_experiments/pck/figures_pck/#1/\Ours/trg_1x_#2_crop.jpg}\\
}
 

\newcommand{\QualiRowC}[3]{
    \inci{figures/05_experiments/pck/figures_pck/#1/dinov2_b14_518_upft1/src_10_#2_crop.jpg}
    &\inci{figures/05_experiments/pck/figures_pck/#1/dinov2_b14_518_upft1/trg_10_#2_gt_crop.jpg}
    &\inci{figures/05_experiments/pck/figures_pck/#1/dinov2_b14_518_upft1/trg_10_#2_crop.jpg}
    &\inci{figures/05_experiments/pck/figures_pck/#1/\Geo/trg_10_#2_crop.jpg}
    &\inci{figures/05_experiments/pck/figures_pck/#1/\Ours/trg_10_#2_crop.jpg}\\
}
\newcommand{\QualiRowCCUB}[3]{
    \inci{figures/05_experiments/pck/figures_pck/#1/dinov2_b14_518_upft1/src_1x_#2_crop.jpg}
    &\inci{figures/05_experiments/pck/figures_pck/#1/dinov2_b14_518_upft1/trg_1x_#2_gt_crop.jpg}
    &\inci{figures/05_experiments/pck/figures_pck/#1/dinov2_b14_518_upft1/trg_1x_#2_crop.jpg}
    &\inci{figures/05_experiments/pck/figures_pck/#1/\Geo/trg_1x_#2_crop.jpg}
    &\inci{figures/05_experiments/pck/figures_pck/#1/\Ours/trg_1x_#2_crop.jpg}\\
}

\begin{figure*}[ht!]
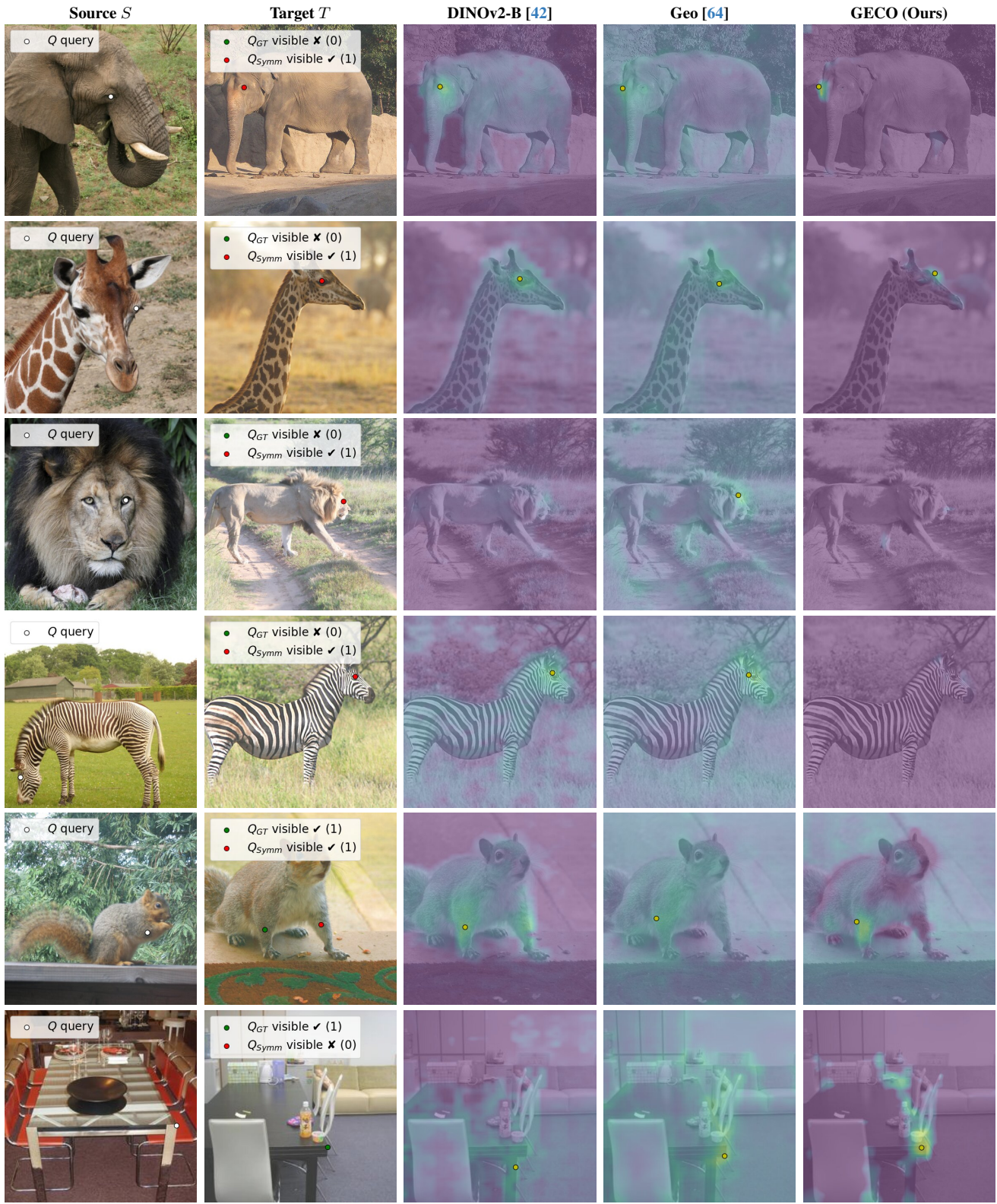

    \centering
    \footnotesize
    \begin{tabularx}{\linewidth}{*{5}{Y}}
      \textbf{Source $S$}
      &\tabularxmulticolumncentered{1}{c}{\textbf{Target $T$}}
      &\textbf{DINOv2-B \cite{oquab2023dinov2}}
      &\textbf{Geo \cite{zhang2024telling}}
      &\textbf{\methodname (Ours)}\\
      \QualiRowA{elephant}{5}{ap10kpairs}
      \QualiRowA{giraffe}{5}{ap10kpairs}
      \QualiRowA{lion}{13}{ap10kpairs}
      \QualiRowA{zebra}{7}{ap10kpairs}
      \QualiRowB{squirrel}{11}{ap10kpairs}
      \QualiRowBPF{table}{14}{pfpascalpairs}
    \end{tabularx}
    \caption{\textbf{Qualitative results on correspondence estimation for APK~\cite{zhang2024telling}, and PFPascal~\cite{ham2017proposal}.}
    (Top two rows) The model accurately locates the ground truth correspondence, becoming visible with slight movement, while ignoring symmetric counterparts.
    (Third, Fourth row) When the symmetric counterpart is occluded, attention is uniformly low across the image.}\label{fig:pck_quanti1}
\end{figure*}

\begin{figure*}[ht!]
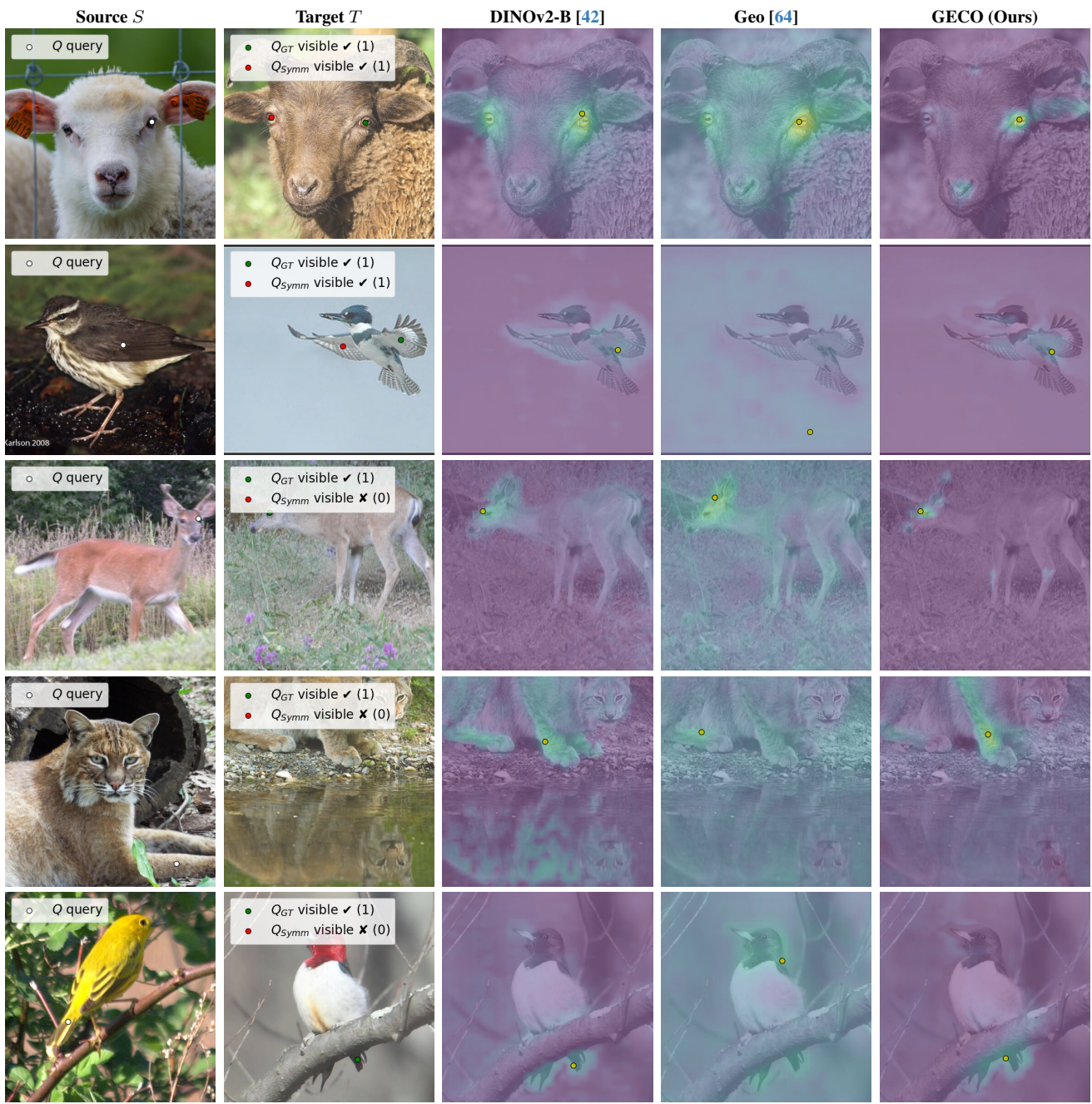

    \centering
    \footnotesize
    \begin{tabularx}{\linewidth}{*{5}{Y}}
      \textbf{Source $S$}
      &\tabularxmulticolumncentered{1}{c}{\textbf{Target $T$}}
      &\textbf{DINOv2-B \cite{oquab2023dinov2}}
      &\textbf{Geo \cite{zhang2024telling}}
      &\textbf{\methodname (Ours)}\\
      \QualiRowB{sheep}{6}{ap10kpairs}
      \QualiRowBCUB{bird}{2}{CUB_200_2011}
      \QualiRowC{deer}{6}{ap10kpairs}
      \QualiRowC{bobcat}{2}{ap10kpairs}
      \QualiRowCCUB{bird}{7}{CUB_200_2011}
    \end{tabularx}
    \caption{\textbf{Qualitative results on correspondence estimation for APK~\cite{zhang2024telling}, and CUB~\cite{wah2011caltech}.}
    (Bottom row) CUB samples confirm that our method preserves the pretrained foundation model’s generalization.}\label{fig:pck_quanti2}
\end{figure*}

\begin{table*}[ht!]
  \footnotesize
  \centering
      \begin{tabularx}{\linewidth}{m{1.9cm} *{2}{Y} *{2}{Y} *{2}{Y} *{2}{Y}}
        \hline
        & \tabularxmulticolumncentered{2}{c}{\textbf{PFPascal~\cite{ham2017proposal}}}
        & \tabularxmulticolumncentered{2}{c}{\textbf{APK~\cite{zhang2024telling}}} 
        & \tabularxmulticolumncentered{2}{c}{\textbf{Spair~\cite{min2019spair}}} 
        & \tabularxmulticolumncentered{2}{c}{{\color{blue}\textbf{CUB~\cite{wah2011caltech}}}} \\
         \cmidrule(lr){2-3} \cmidrule(lr){4-5} \cmidrule(lr){6-7} \cmidrule(lr){8-9}\\
        &PCK$\uparrow$ & time[ms]$\downarrow$ 
        &PCK$\uparrow$ & time[ms]$\downarrow$
        &PCK$\uparrow$ & time[ms]$\downarrow$
        &PCK$\uparrow$ & time[ms]$\downarrow$\\
        \hline
        DINO~\cite{caron2021emerging} 
        & 65.3 & 26 
        & 51.8 & 22
        & 48.4 & 25
        & 75.6 & 26\\
        DIFT$_{\text{ad}}$~\cite{tang2023emergent}
        & 72.5 & 221  
        & 60.4 & 228
        & 59.3 & 222
        & 84.2 & 222\\
        DINOv2-S~\cite{oquab2023dinov2} 
        & 85.5 & \textbf{14}  
        & 71.7 & \textbf{14}
        & 66.3 & \textbf{15}
        & 92.4 & \textbf{17}\\
        DINOv2-B~\cite{oquab2023dinov2} 
        & 86.0 & {45}  
        & 73.0 & {40}
        & 67.1 & {38}
        & \textbf{92.8} & {43}\\
        \hline
        Geo~\cite{zhang2024telling}
        &86.1& 2141
        &80.5& 2127
        &\textbf{90.1}& 2159
        &88.4& 2274\\ 
        Sphere~\cite{mariotti2024improving}
        &88.5&2152
        &75.2&2144
        &74.5&2164
        &92.1&2151\\
        \hline
        \methodname (Ours)
        & {\textbf{92.1}} & {45}
        & {\textbf{86.7}} & {40}
        & {85.2} & {38}
        & {\textbf{92.5}} & {43} \\
        \hline
     \end{tabularx}
  \caption{\textbf{Quantitative evaluation of PCK on PFPascal~\cite{ham2017proposal}~(Left), APK~\cite{zhang2024telling}~(Middle), Spair~~\cite{min2019spair}~(Middle), and CUB~\cite{wah2011caltech}~(Right).} We report the PCK@$\alpha=0.1$ for four different datasets on the test split. Our method outperforms competitors in three out of four datasets by 6.0\% on PFPascal, 6.2\% on APK, and 4.1\% on CUB, while $\sim$2 orders of magnitude faster. In contrast, does not generalize well to CUB, where it lags behind DINOv2-S. Both Geo and our method are trained on PFPascal~\cite{ham2017proposal}, APK~\cite{zhang2024telling}, and Spair~\cite{min2019spair}, while Sphere is trained solely on Spair. The best scores are highlighted in \textbf{bold}. Methods are considered to be on par if their performance difference is less than 0.5\%. We mark {\color{blue}\textbf{CUB~\cite{wah2011caltech}}} in blue to highlight that it has not been seen at training time. }\label{tab:pck_pfpascal_apk_spair}
\end{table*}

\subsubsection{Evaluation}

\paragraph{Data}
In this experiment, we evaluate our method on datasets with pairwise keypoint annotations. Specifically, we use the CUB dataset~\cite{wah2011caltech}, selecting 10,000 image pairs of at random, as well as SPair~\cite{min2019spair}, PFPascal~\cite{ham2017proposal}, and APK~\cite{zhang2024telling}, which provide predefined image pairs. Since PFPascal lacks symmetric counterpart annotations, we assess only the standard PCK metric on this dataset.

\paragraph{Results}
The quantitative PCK analysis on PFPascal, APK, and Spair is shown in~\cref{tab:pck_pfpascal_apk_spair}. 
Our method surpasses previous state-of-the-art by 6.0\% on PFPascal, 6.2\% on APK, and 4.1\% on CUB, while being nearly two orders of magnitude faster. Notably, Geo, using DINOv2-B and Stable Diffusion features, generalizes less well on CUB than DINOv2-S. In fact, the only dataset where Geo outperforms our method is Spair, indicating that is not a general purpose model.
Detailed PCK results for PFPascal, APK, and Spair appear in the supplementary material. We provide an in-depth APK evaluation in \cref{tab:pgck_splits}, where our method improves across all $n_{11}$ splits and consistently outperforms Geo in three of four cases (see supplementary).
\subsection{Feature Space Segmentation}\label{sec:experiments:segmentation}
The PCK metric evaluates correspondence accuracy based on a limited set of keypoints. While effective at those points, it does not assess the dense feature space, which is crucial since methods are trained on these annotations. We propose evaluating the feature space by partitioning it into semantically meaningful parts using only Euclidean distance to part-specific representation vectors. This indicates whether the learned features capture meaningful, consistent structures within the image.

\subsubsection{Evaluation}\label{sec:experiments:05_experiments_02_segmentation}

\paragraph{Centroid Clustering} To analyze the feature-space structure, we compute centroids for each annotated object part, assessing whether the model can separate parts using only Euclidean distance to these centroids. We first gather sets of feature vectors corresponding to each annotated part and then compute their median as the part representations. On the test set, each patch is assigned to the nearest centroid based on Euclidean distance, evaluating the model’s ability to discriminate parts purely from the learned features.

\paragraph{Data}
We evaluate dense features using part annotations from PascalParts~\cite{chen2014detect}, which offer consistent, category-specific labels for assessment.

\paragraph{Results}
Our learned feature representations reliably distinguish semantically similar parts with distinct geometric properties. \cref{fig:clustering1,fig:clustering2} show qualitative examples where our method accurately separates challenging regions such as left/right eyes, wings, and ears. In contrast, the foundation model often exhibits artifacts, marked by red arrows and circles in the figure. Geo~\cite{zhang2024telling}, which uses Gaussian sampling around keypoints during training, tends to assign overly broad regions to the eyes. Quantitative and qualitative confusion matrix analyses in the supplementary highlight geometric ambiguities in foundation models. Our method achieves geometrical awareness comparable to Geo~\cite{zhang2024telling} and shows greater consistency on non-geometric parts.
\newcommand{\QualitativeRowSeg}[3]{
    \inci{figures/05_experiments/segmentation/#3/#1/GT/#2_RGB.jpg}
    &\inci{figures/05_experiments/segmentation/#3/#1/dinov2_b14_518_upft1/#2.jpg}
    &\inci{figures/05_experiments/segmentation/#3/#1/\Geo/#2.jpg}
    &\inci{figures/05_experiments/segmentation/#3/#1/\Ours/#2.jpg}
    &\inci{figures/05_experiments/segmentation/#3/#1/GT/#2.jpg}\\
} 

\begin{figure*}[ht!]
    \centering
    \footnotesize
    \begin{tabularx}{\linewidth}{*{5}{Y}}
      \textbf{Input}
      &\textbf{DINOv2-B \cite{oquab2023dinov2}}
      &\textbf{Geo \cite{zhang2024telling}}
      &\textbf{\methodname (Ours)} 
      &\textbf{Ground Truth}\\
      \QualitativeRowSeg{bird}{14}{pascalparts}
      \QualitativeRowSeg{bird}{72}{pascalparts}
      \QualitativeRowSeg{bird}{82}{pascalparts}
      \QualitativeRowSeg{bird}{125}{pascalparts}
      \QualitativeRowSeg{bird}{128}{pascalparts}
      \multicolumn{5}{c}{
      \begin{overpic}[trim=0cm 0.0cm 0cm 0.0cm,clip, height=0.017\linewidth]{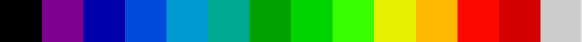}
      \put(0, -17){\rotatebox{90}{bkg}}
      \put(7.14, -17){\rotatebox{90}{head}}
      \put(14.28, -17){\rotatebox{90}{leye}}
      \put(21.43, -17){\rotatebox{90}{reye}}
      \put(28.57, -17){\rotatebox{90}{beak}}
      \put(35.71, -17){\rotatebox{90}{torso}}
      \put(42.85, -17){\rotatebox{90}{neck}}
      \put(50.00, -17){\rotatebox{90}{lwing}}
      \put(57.14, -17){\rotatebox{90}{rwing}}
      \put(64.29, -17){\rotatebox{90}{lleg}}
      \put(71.43, -17){\rotatebox{90}{lfoot}}
      \put(78.57, -17){\rotatebox{90}{rleg}}
      \put(85.71, -17){\rotatebox{90}{rfoot}}
      \put(92.85, -17){\rotatebox{90}{tail}}
      \end{overpic}
      }\\
      \vspace{1.3cm}\\
    \end{tabularx}
    \caption{\textbf{Clustering of the feature space based on Euclidean distance to a part representation vector for PascalParts~\cite{chen2014detect}.}
    Our learned representation effectively separates even challenging parts, such as left and right eyes, wings, and ears, while also being similarly time and memory efficient as the DINOv2v2-B backbone and much more time efficient than Geo.}\label{fig:clustering1}
\end{figure*}
\begin{figure*}[ht!]
  \centering
  \footnotesize
  \begin{tabularx}{\linewidth}{*{5}{Y}}
    \textbf{Input}
    &\textbf{DINOv2-B \cite{oquab2023dinov2}}
    &\textbf{Geo \cite{zhang2024telling}}
    &\textbf{\methodname (Ours)} 
    &\textbf{Ground Truth}\\
    \QualitativeRowSeg{cat}{0}{pascalparts}
    \QualitativeRowSeg{cat}{277}{pascalparts}
    \QualitativeRowSeg{cat}{356}{pascalparts}
    \QualitativeRowSeg{cat}{696}{pascalparts}
    \QualitativeRowSeg{cat}{714}{pascalparts}
    \multicolumn{5}{c}{
    \begin{overpic}[trim=0cm 0.0cm 0cm 0.0cm, clip,height=0.017\linewidth]{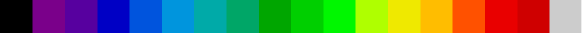}
      \put(0.000000, -12.2){\rotatebox{90}{bkg}}
      \put(5.555556, -12.2){\rotatebox{90}{head}}
      \put(11.111111, -12.2){\rotatebox{90}{leye}}
      \put(16.666667, -12.2){\rotatebox{90}{reye}}
      \put(22.222222, -12.2){\rotatebox{90}{lear}}
      \put(27.777778, -12.2){\rotatebox{90}{rear}}
      \put(33.333333, -12.2){\rotatebox{90}{nose}}
      \put(38.888889, -12.2){\rotatebox{90}{torso}}
      \put(44.444444, -12.2){\rotatebox{90}{neck}}
      \put(50.000000, -12.2){\rotatebox{90}{lfleg}}
      \put(55.555556, -12.2){\rotatebox{90}{lfpa}}
      \put(61.111111, -12.2){\rotatebox{90}{rfleg}}
      \put(66.666667, -12.2){\rotatebox{90}{rfpa}}
      \put(72.222222, -12.2){\rotatebox{90}{lbleg}}
      \put(77.777778, -12.2){\rotatebox{90}{lbpa}}
      \put(83.333333, -12.2){\rotatebox{90}{rbleg}}
      \put(88.888889, -12.2){\rotatebox{90}{ebpa}}
      \put(94.444444, -12.2){\rotatebox{90}{tail}}
    \end{overpic}
    }\\
    \vspace{1.3cm}\\
  \end{tabularx}
  \caption{\textbf{Clustering of the feature space based on Euclidean distance to a part representation vector for PascalParts~\cite{chen2014detect}.}
  Our learned representation effectively separates even challenging parts, such as left and right eyes, wings, and ears, while also being similarly time and memory efficient as the DINOv2v2-B backbone and much more time efficient than Geo.}\label{fig:clustering2}
\end{figure*}
\subsection{Runtime}\label{sec:experiments:runtime}

\paragraph{Procedure}
We measure inference time by running a forward pass (excluding image loading) on 1,000 images per dataset using an RTX A4000 GPU, averaging the results. All models are evaluated with a batch size of 1.

\paragraph{Results}
Our timing results can be found in \cref{tab:pck_pfpascal_apk_spair}. For DINOv2 our measurements are consistent with those reported in the NVIDIA NCG catalog~\cite{nvidia2023ngc}, falling within the same order of magnitude. The addition of our ``lightspeed" adapter introduces minimal overhead, contributing less than $0.5$ ms to DINOv2-B's baseline runtime of approximately $40$ ms. Importantly, the performance of the original DINOv2 on the geometric matching task is not as strong. On the other hand, Geo's~\cite{zhang2024telling} geometrically aware features, which rely on diffusion models, result in much longer inference times exceeding $2$ seconds. Furthermore, in contrast to Geo~\cite{zhang2024telling}, our method benefits from lower memory requirements, enabling efficient batch processing at inference time, which would further amplify the speed performance gap, making our method significantly faster and more scalable for practical applications.
\section{Conclusion}
We present a fast and efficient representation learning method based on optimal transport loss, achieving state-of-the-art PCK performance and improved geometric understanding. Our structured analysis highlights underexplored aspects of feature learning using PCK subdivisions and centroid clustering. While our method is lightweight and generalizes well, it relies on sparse keypoint supervision and is currently limited to categories with available annotations.

\clearpage

{\footnotesize
\noindent
\textbf{Acknowledgements:} 
We thank the anonymous reviewers for their valuable feedback. This work was supported by the ERC Advanced Grant SIMULACRON, cby the Federal Ministry for the Environment, Nature Conservation, Nuclear Safety and Consumer Protection (BMUV) through the AuSeSol-AI project (grant 67KI21007A), and by the TUM Georg Nemetschek Institute Artificial Intelligence for the Built World (GNI) through the AICC project.
}
{
    \small
    \bibliographystyle{ieeenat_fullname}
    \bibliography{main}
}

\maketitlesupplementary
\tableofcontents

\section{Implementation Details}

\subsection{Architecture Details}

\paragraph{Feature Encoder Hyperparameters}
We identified notable differences in the optimal settings for DINOv2 and DINO that are not explicitly documented in their original papers. For DINOv2, we achieve the best performance by using patch tokens from the final transformer layer, particularly when working with higher-resolution images. In contrast, DINO performs better when using patch tokens from the fourth-to-last transformer layer, with optimal results at an input resolution of 244; performance degrades with higher resolutions.

Based on these findings, we use the DINOv2 backbone~\cite{oquab2023dinov2} with an input resolution of 518 and patch tokens from the final transformer layer in our model.

\paragraph{Architecture}

We learn low-rank matrices (rank 10) for each attention layer, which are added to the \( 768 \times 768 \)-dimensional weight matrices of the linear layers before being passed to the attention mechanism (\( 768 \times 10 \times 2\) parameters). For DINOv2-B, this results in \( v = 768 \times 10 \times 2 \times 12 \times 2\) parameters, with a depth of 12 and  updating query and value matrices equating \( v \times 4 \) bytes (\(\approx 1.4\) MB) of learnable parameters. This is significantly fewer than the 19 MB of learnable parameters introduced by Geo~\cite{zhang2024telling}.
This small parameter count and simple architecture enable us to increase the inference time of DINOv2-B by less than 1 millisecond.

\subsection{Training Details}

\paragraph{Training Hyperparameters}

The models are trained using the Adam optimizer with a learning rate of 0.0001, no weight decay, and no learning rate scheduler, with a batch size of 6. A key advantage of our approach is the ability to train directly on raw images without preprocessing, which allows for increased data augmentation and larger batch sizes. Training is conducted on a single GPU for 8 epochs. For the loss functions, weights are set as follows: 1 for the positive loss, 1 for the bin loss, and 10 for the negative loss.

\paragraph{Dataset Shuffeling}
Training is conducted jointly across the PFPascal, SPair~\cite{min2019spair}, and APK~\cite{zhang2024telling} datasets. We employ a reweighting scheme that samples up to 800 examples per category. We found this hyperparameter by trading off the validation accuracies for the different datasets (see~\cref{tab:ablation_study_dataset_shuffling}).
In fact, if one dataset dominates the training, the model tends to overfit to that dataset.

\paragraph{Image Pair Augmentation}
We leverage existing datasets~\cite{wah2011caltech,ham2017proposal,zhang2024telling,min2019spair} containing images of diverse category instances with varying shapes, textures, and motion deformations, captured under different lighting and camera setups. Thanks to our method's low computational cost, we use more data augmentation than~\cite{zhang2024telling} to improve generalization.
We incorporate flipping, cropping, and color jitter into the augmentation pipeline, generating more positive, negative, and bin pairs for feature matching (see \cref{fig:architecture}). Specifically, flipping creates challenging positive/negative pairs, while cropping increases bin pairs by promoting matches to the bin rather than to semantically similar but geometrically inconsistent features.
\begin{figure*}[ht!]
  \centering
  \vspace{-0.0cm}
  \begin{overpic}[angle=0,trim= 0.0cm 0.0cm 0cm 0.0cm,clip, width=0.9\linewidth]{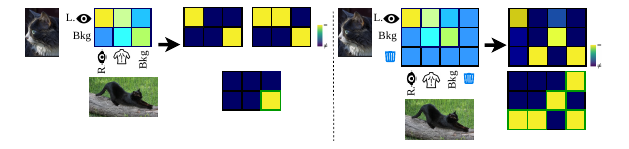}
      \scriptsize
      \put(31,24.5){ArgMax}
      \put(32,23){(rows)}
  
      \put(42,24.5){ArgMax}
      \put(43,23){(cols)}
  
      \put(32,7){GT}
      \put(3,24){\large{\underline{\textbf{ArgMax}}}}
      \put(54,24){\large{\underline{\textbf{Optimal Transport}}}}
  
      \put(38,1.5){\textbf{\textcolor{red}{Sparse}}}
      \put(30,0){\textbf{\textcolor{red}{Information only on Background}}}
  
      \put(78,7){GT}
      \put(78,19){OT}
      \put(87,1.5){\textbf{\textcolor{ForestGreen}{Dense}}}
      \put(80,0){\textbf{\textcolor{ForestGreen}{Information on (dis)occlusions}}}
  \end{overpic}
  \caption{\textbf{Intuition on OT and geometry.} Standard ArgMax loss (left) provides positive supervision only for elements that appear in both images. Optimal Transport (right) uses the bin to provide explicit signals also on elements that appear in only one of the images. This is particularly common for non-rigid shapes (e.g., animals), as often a point and its symmetrical counterpart (e.g., left and right eyes) face (dis)occlusions. Intuitively, this leads to a stronger signal for learning features that better disambiguate symmetries.}
  \label{fig:moticationOT}
\end{figure*}

\subsection{Loss Function}
\paragraph{Motivation for OT}
We argue that Optimal Transport (OT) provides stronger geometric supervision than Argmax, as shown in Fig.~\ref{fig:moticationOT}.
As already pointed out in \cref{sec:background:argmax_matching} there are several drawbacks when using Argmax to construct a loss function. Here, we highlight, that OT incorporates the supervision signal of points, visible in only one image, which often occurs with symmetric parts of non-rigid objects (e.g., eyes) by introducing a dustbin entry in the assignment matrix.
Furthermore, due to the interaction between features, the OT loss enables to densely backpropagte the gradient. For example, a bin assignment signal pushes down the similarity to all other features, which is not possible with Argmax.

\begin{figure}[t!]
  \centering
  \includegraphics[width=1\linewidth]{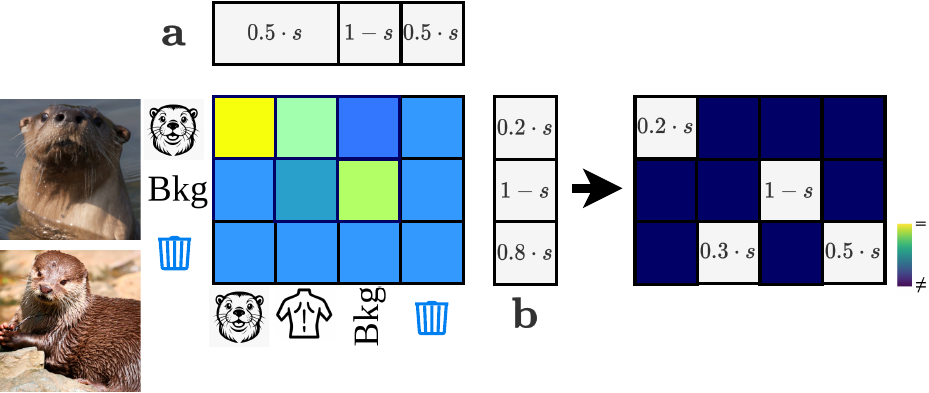}
  \caption{\textbf{Optimal Transport Marginals.} This figure illustrates the marginal distribution used in the Optimal Transport algorithm, assigning $s$ to the shape and $1-s$ to the background. The visible content proportion is estimated from keypoint annotations by calculating the ratio of visible to total keypoints. In this example, the top image has a visibility ratio of 0.2 (only the head is visible), while the bottom image has a ratio of 0.5 (the head and one side of the body are visible) relative to the full shape.
  } \label{fig:ot_marginals}
\end{figure}
\paragraph{Designing marginal distributions}
Since correspondence data is valuable, a key contribution of our work is using it to construct marginal distributions see \cref{fig:ot_marginals}. 
To define the marginals for the Optimal Transport (OT) problem, we depart from the conventional approach, employed in SuperGlue~\cite{sarlin2020superglue}, which assigns 0.9 to the image and 0.1 to the bin. Instead, we utilize automatically generated mask annotations to distinguish between foreground and background, leveraging an off-the-shelf segmentation tool~[\hyperlink{bib:kirillov2023segment}{67}].

In our formulation, the shape mass is assigned a value of \( s = 0.9 \), with the remaining \( 1 - s \) allocated to the background, as illustrated in \cref{fig:ot_marginals}. The proportion of visible content, denoted as \( x \), is estimated by calculating the ratio of visible keypoints to the total number of keypoints. This leads to $x\cdot s$ being assigned to the foreground and $(1-x)\cdot s$ to the bin.

This marginal distribution, together with the padded cosine similarity matrix of the features, is then fed into the Optimal Transport algorithm to determine the optimal transport plan.

\subsection{Ablation Study on Design Choices}

\paragraph{Impact of model size on performance (PCK)}
When scaling down the architecture to DINOv2-S, we observe a decline in performance, which we attribute to the reduced capacity of the model. While it still captures geometric relationships between keypoints, its performance does not match that of the larger variant, as shown in \cref{tab:ablation_study_model_size}.
\begin{table}[t!]
  \footnotesize
  \centering
  \begin{tabularx}{1\linewidth}{m{1.9cm} *{4}{Y}}
    & \tabularxmulticolumncentered{1}{c|}{\textbf{PFPascal~\cite{ham2017proposal}}}
    & \tabularxmulticolumncentered{1}{c|}{\textbf{APK~\cite{zhang2024telling}}} 
    & \tabularxmulticolumncentered{1}{c|}{\textbf{Spair~\cite{min2019spair}}} 
    & \tabularxmulticolumncentered{1}{c}{{\color{blue}\textbf{CUB~\cite{wah2011caltech}}}} \\
    \hline
    \methodname (Ours)-S & 89.6 & 83.4 & 78.0 & 90.4 \\
    \methodname (Ours) & \textbf{92.1} & \textbf{86.7} & \textbf{85.2} & \textbf{92.5} \\
  \end{tabularx}
  \caption{\textbf{Impact of model size on performance (PCK~$\uparrow$).} Comparison of our approach using DINOv2-B and DINOv2-S backbones on PFPascall~\cite{ham2017proposal}, CUB~\cite{wah2011caltech}, SPair~\cite{min2019spair}, and APK~\cite{zhang2024telling} datasets. The results indicate that model size plays a crucial role in overall performance.} \label{tab:ablation_study_model_size}
\end{table}

\paragraph{Impact of the Optimal Transport Marginals}
We evaluate the impact of the marginal distribution on the performance of our method. We compare the standard marginals used in SuperGlue~\cite{sarlin2020superglue} with our approach, which leverages mask annotations to distinguish between foreground and background. We found that our approach outperforms the standard marginals especially on the generalization to the CUB~\cite{wah2011caltech} dataset, as shown in \cref{tab:ablation_study_marginals}.

\begin{table}[t!]
  \footnotesize
  \centering
  \begin{tabularx}{1\linewidth}{m{1.9cm} *{4}{Y}}
    & \tabularxmulticolumncentered{1}{c|}{\textbf{PFPascal~\cite{ham2017proposal}}}
    & \tabularxmulticolumncentered{1}{c|}{\textbf{APK~\cite{zhang2024telling}}} 
    & \tabularxmulticolumncentered{1}{c|}{\textbf{Spair~\cite{min2019spair}}} 
    & \tabularxmulticolumncentered{1}{c}{{\color{blue}\textbf{CUB~\cite{wah2011caltech}}}} \\
    \hline
    \methodname (Ours)-N & 91.5 & 86.0 & 83.7 & 89.5 \\
    \methodname (Ours) & \textbf{92.1} & \textbf{86.7} & \textbf{85.2} & \textbf{92.5} \\
  \end{tabularx}
  \caption{\textbf{Impact of choice of marginals on performance (PCK~$\uparrow$).} We evaluate the effect of incorporating more sophisticated marginals (Bottom row) in the loss function compared to the standard formulation (Top row). Our results show that while the performance difference on the test splits of APK, PFPascal, and SPair remains within a margin of 1.5, the performance drop on the generalization task is substantially larger, reaching a value of 3. This highlights the importance of incorporating improved marginals for better generalization.}\label{tab:ablation_study_marginals}
\end{table}

\paragraph{Impact of the KL Divergence Regularization on the marginals}
The benefit of KL regularization is clear in the segmentation results, as shown in \cref{tab:ablation_study_kl}.  Because the estimated marginals are imperfect,
\begin{table}[t!]
    \footnotesize
    \centering 
    \begin{tabularx}{1\linewidth}{m{2.3cm} *{2}{Y}}
        &\tabularxmulticolumncentered{1}{c|}{mean mIoU $\uparrow$} & \tabularxmulticolumncentered{1}{c}{mean Acc $\uparrow$} \\
          \hline
          {\methodname} (Ours) \tiny{(w/o KL)}  & 36.4 & 88.5\\
          {\methodname} (Ours) & \textbf{37.9} & \textbf{89.0}\\
    \end{tabularx}
    \caption{\textbf{Impact of choice of marginals on performance (Segmentation Accuracy).} 
    We evaluate using the segmentation metrics explained in \cref{secsup:experiments:05_experiments_02_segmentation}. We show the effect of incorporating the KL-marginal regularization (Bottom row) in the loss function compared to the standard formulation with hard contraints on the marginal distributions (Top row).}\label{tab:ablation_study_kl}
\end{table}
 KL reduces overfitting to noise during training and should be increased when exact marginals are available (\eg, images rendered from 3D shapes).

 \paragraph{Impact of dataset shuffeling on the performance}\label{tab:ablation_study_dataset_shuffling}
 We evaluate the effect of dataset shuffling on performance across different datasets. As our method is trained jointly on SPair~\cite{min2019spair}, PFPascal~\cite{ham2017proposal}, and APK~\cite{zhang2024telling}, shuffling enables better generalization. To address category imbalance, we sample up to $K$ image pairs per category; if fewer are available, all are used.

 Performance is highly sensitive to the choice of $K$. Low values (e.g., 100-400) lead to overfitting on PFPascal and poor generalization to SPair and CUB. Conversely, high $K$ underrepresents PFPascal, limiting its validation accuracy. We find that $K = 800$ yields the best overall performance.
\vspace{3cm}
\section{Additional Experiments}
We provide more detailled experiments to analyze the PCK metric and its subsets in \cref{secsup:experiments:05_experiments_00_pck_analysis}.
In \cref{secsup:experiments:05_experiments_02_segmentation}, we complement the qualitative analysis of the segmentation performance in the main paper with a quantitative and qualitative evaluation of the segmentation confusion matrix.
In \cref{secsup:experiments:05_experiments_03_pnp}, we provide an additional experiment focusing on rigid shapes with diverse appearances and demonstrate our methods performance on keypoint matching by evaluating 2D-3D projections.

\subsection{Analysis with (and of) PCK}\label{secsup:experiments:05_experiments_00_pck_analysis}
\begin{figure*}[ht!]
  \centering
  \begin{subfigure}{0.3\linewidth}
    \centering
    \caption{\centering {\color{mygreen}$\frac{\underline{n}_{11}}{n_{11}}\uparrow$} on {\color{blue}CUB~\cite{wah2011caltech}}}
    \label{fig:pgck_splits_radii_cub_underl}
    \includegraphics[width=1\linewidth]{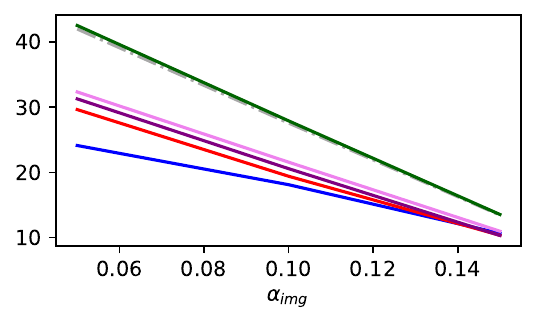}
  \end{subfigure}
  \begin{subfigure}{0.3\linewidth}
    \centering
    \caption{\centering {\color{mygreen}$\frac{\underline{n}_{11}}{n_{11}}\uparrow$} on APK~\cite{zhang2024telling}}
    \label{fig:pgck_splits_radii_apk_underl}
    \includegraphics[width=1\linewidth]{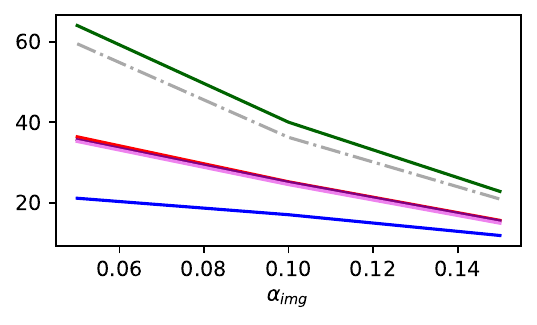}
  \end{subfigure}
  \begin{subfigure}{0.3\linewidth}
    \centering
    \caption{\centering {\color{mygreen}$\frac{\underline{n}_{11}}{n_{11}}\uparrow$} on Spair~\cite{min2019spair}}
    \label{fig:pgck_splits_radii_spair_underl}
    \begin{overpic}[trim=0cm 0cm 0cm 0cm,clip,width=1.0\linewidth]{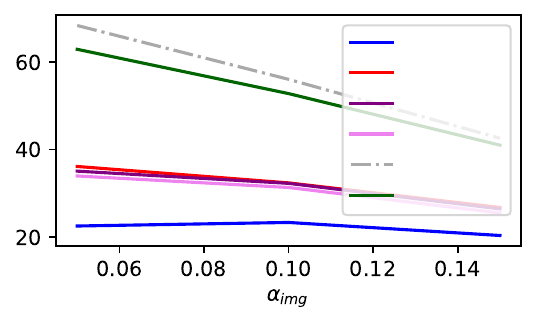}
        \put(74,51.3){\tiny DINO~\cite{caron2021emerging}}
        \put(74,45.8){\tiny DIFT~\cite{tang2023emergent}}
        \put(74,40.3){\tiny DINOv2S~\cite{oquab2023dinov2}}
        \put(74,34.8){\tiny DINOv2B~\cite{oquab2023dinov2}}
        \put(74,29.3){\tiny Geo~\cite{zhang2024telling}}
        \put(74,23.8){\tiny  \textbf{\methodname} \textbf{(Ours)}}
    \end{overpic}
  \end{subfigure}
  \caption{\textbf{Geometric ambiguity as a function of radius.}  As the radius decreases, the set of unambiguous true positives {\color{mygreen}$\frac{\underline{n}_{11}}{n_{11}}$} grows, where the target keypoint is outside the radius of any incorrect matches. This figure illustrates the PGCK subset, specifically the unambiguous correct predictions, for various radius values (@$\alpha_{img}$) on CUB~\cite{wah2011caltech}, APK~\cite{zhang2024telling}, and Spair~\cite{min2019spair}. Our method outperforms previous work on two out of three datasets while achieving a reduction in runtime and memory usage by two orders of magnitude.}\label{fig:pgck_splits_radii}
\end{figure*}
\begin{figure}[ht!]
  \centering
  \begin{subfigure}{0.66\linewidth}
    \centering
    \caption{\centering{PCK$\uparrow$} on PFPascal~\cite{ham2017proposal}}
    \label{fig:pgck_splits_radii_pfpascal_all}
    \includegraphics[width=1\linewidth]{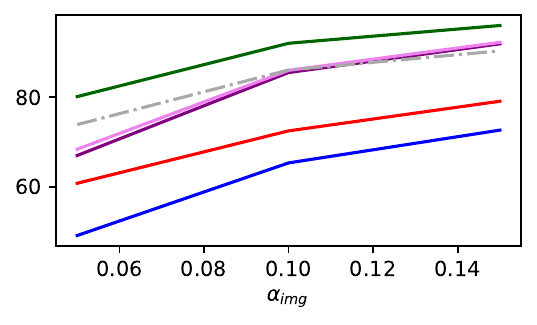}
  \end{subfigure}
  \begin{subfigure}{0.66\linewidth}
    \centering
    \caption{\centering{PCK$\uparrow$} on Spair~\cite{min2019spair}}
    \label{fig:pck_radii_spair_all}
    \includegraphics[width=1\linewidth]{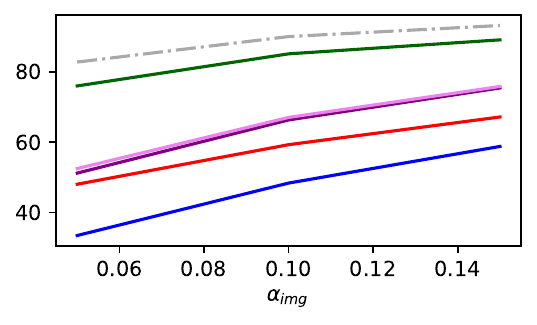}
  \end{subfigure}\\
  \begin{subfigure}{0.66\linewidth}
    \centering
    \caption{\centering{PCK$\uparrow$} on {\color{blue}CUB~\cite{wah2011caltech}}}
    \label{fig:pck_radii_cub_all}
    \includegraphics[width=1\linewidth]{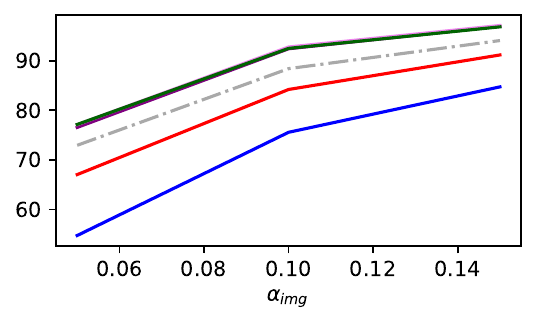}
  \end{subfigure}
  \begin{subfigure}{0.66\linewidth}
    \centering
    \caption{\centering{PCK$\uparrow$} on APK~\cite{zhang2024telling}}
    \label{fig:pck_splits_apk_all}
    \begin{overpic}[trim=0cm 0cm 0cm 0cm,clip,width=1.0\linewidth]{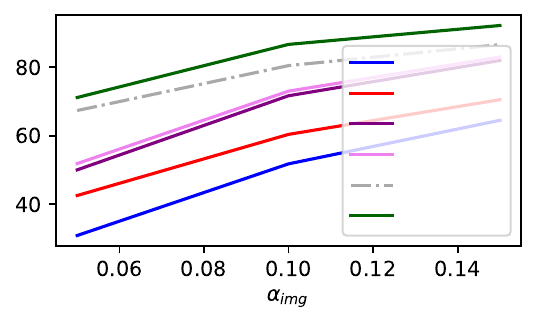}
      \put(74,47.3){\tiny DINO~\cite{caron2021emerging}}
      \put(74,41.8){\tiny DIFT~\cite{tang2023emergent}}
      \put(74,36.3){\tiny DINOv2S~\cite{oquab2023dinov2}}
      \put(74,30.8){\tiny DINOv2B~\cite{oquab2023dinov2}}
      \put(74,25.3){\tiny Geo~\cite{zhang2024telling}}
      \put(74,19.5){\tiny \methodname (Ours)}
    \end{overpic}
  \end{subfigure}
  \caption{\textbf{PCK as a function of radius.} As the radius decreases, the set of correct matches declines. This figure illustrates the PCK metric for various radius values (@$\alpha_{img}$) on CUB~\cite{wah2011caltech}, APK~\cite{zhang2024telling}, and Spair~\cite{min2019spair}. Our method outperforms previous work on three out of four datasets while achieving a reduction in runtime and memory usage by two orders of magnitude.}\label{fig:pck_radii}
\end{figure}

\paragraph*{Influence of different radii on unambiguous True Positives}
A critical choice of PCK is the radius in which a prediction is considered correct. This also has an impact on our PGCK subdivision. Here, we investigate how the choice of the PCK metric's radius influences the unambiguous true positive subset's cardinality (see \cref{fig:pgck_splits_radii}). As expected, the unambiguous correct predictions decrease with the radius, while the overall PCK increases (see \cref{fig:pck_radii}).
For larger radius values, the PCK metric reflects the model's ability to identify general correspondences across the image, even when semantic parts are widely distributed rather than concentrated in a single area. In contrast, for smaller radius values, the metric emphasizes the model's precision in pinpointing correspondences with high spatial accuracy and its ability to distinguish between closely spaced keypoints, such as those on the left and right sides, when both are present.
\paragraph{Results}
In the main paper, we reported the subsets of the PGCK for one value of \cref{fig:pgck_splits_radii} on APK~\cite{zhang2024telling}. Here, we extend this analysis to the CUB~\cite{wah2011caltech} and SPair~\cite{min2019spair} datasets.
As shown in \cref{fig:pgck_splits_radii_cub_underl}, our method achieves a slight improvement over previous work in unambiguous correct predictions on the CUB~\cite{wah2011caltech} dataset. For the APK~\cite{zhang2024telling} and SPair~\cite{min2019spair} datasets, performance varies: our method outperforms the competitor in one case, while in the other, the competitor demonstrates a more refined geometric understanding. These results suggest that geometric understanding is dataset-dependent.

Additionally, given the performance drop on CUB observed in competing methods on the semantic understanding task (see \cref{fig:pck_radii_cub_all}), we argue that our approach strikes a favorable balance between geometric and semantic understanding. Unlike the competitors, our method preserves previously learned properties, making it a more robust and well-rounded solution. Moreover, it remains competitive with the state of the art while significantly reducing memory consumption and runtime from 2274 ms to 43 ms.

\paragraph*{Influence of different radii on PCK}
As expected the PCK values are increasing with varying radius. It measures the general semantic knowledge of features with a part also measuring the geometric understanding.
\paragraph{Results}

We present detailed results, evaluated on the PFPascal~\cite{ham2017proposal}, APK~\cite{zhang2024telling}, SPair~\cite{min2019spair}, and CUB~\cite{wah2011caltech} datasets. Specifically, in \cref{fig:pck_radii} we show the PCK values for different radii across these datasets. Notably, our method significantly outperforms the current state-of-the-art on three out of the four datasets, highlighting the superior effectiveness of our approach. While the competitor Geo~\cite{zhang2024telling} achieves comparable performance to ours in geometric understanding on the generalization task to CUB~\cite{wah2011caltech} (see \cref{fig:pgck_splits_radii_cub_underl}), we observe significant catastrophic forgetting in the $n_{10}$ and $n_{1x}$ splits leading to worse overall results than DINOv2-S (see \cref{fig:pck_radii_cub_all}). In contrast, our method maintains consistent performance across all splits and outperforms Geo~\cite{zhang2024telling} in three out of four cases, demonstrating its robustness and reliability.

\begin{figure}[ht!]
    \centering
    \footnotesize
    \centering
    \footnotesize
    \setlength{\tabcolsep}{0pt}
      \makebox[\linewidth][c]{%
        \begin{tabular}{*{4}{C{0.25\linewidth}}}
          \textbf{Source $S$} &
          \textbf{Target $T$} &
          \textbf{Geo~\cite{zhang2024telling}} &
          \textbf{\methodname~(Ours)} \\
        \end{tabular}
       } \\
      \includegraphics[valign=c, width=\linewidth]{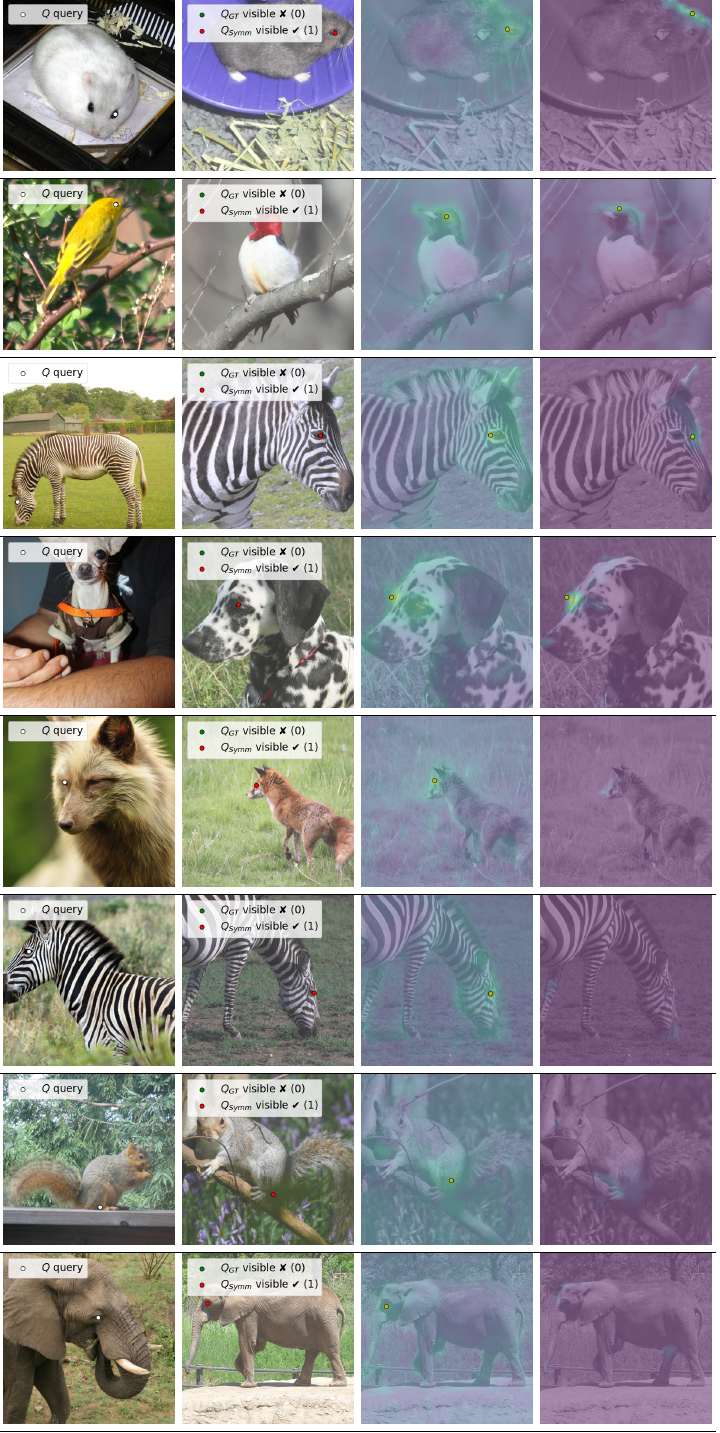} 
      \caption{\textbf{Qualitative results on the task of assignment to the bin (01-case) on APK~\cite{zhang2024telling} and CUB~\cite{wah2011caltech} (fifth row) .} We show that our method predicts small cosine similarities, if the actual match is occluded. This is learned by the bin and negative losses, which encourage the assignment to the bin instead of the symmetric counterpart.
    (First five rows) It is clearly visible, that the model knows the location of the ground truth correspondence, which could become visible by a small movement. The location of the symmetric counterpart is ignored.
    (Last three rows) We receive only low attention values for the whole image, including the symmetric counterpart, when the symmetric counterpart is occluded.}\label{fig:pck_01_failure}
\end{figure}

\paragraph*{Qualitative results}
We illustrate various failure modes of existing approaches and demonstrate how our method addresses these issues. In \cref{fig:pck_01_failure}, we show that our method can correctly assign a keypoint to the bin even when the actual match is occluded. This ability is learned through bin and negative losses, which encourage the assignment to the bin rather than to the symmetric counterpart. When the match is occluded, the model becomes better calibrated, generating minimal attention on the target image, including the symmetric counterpart. This represents a failure mode that the PCK metric does not capture, as the symmetric counterpart is absent in the target image, and consequently, this keypoint pair is excluded from the PCK evaluation.

\begin{figure}[ht!]
  \centering
  \footnotesize
  \setlength{\tabcolsep}{0pt}
  \makebox[\linewidth][c]{%
    \begin{tabular}{*{4}{C{0.25\linewidth}}}
      \textbf{Source $S$} &
      \textbf{Target $T$} &
      \textbf{Geo~\cite{zhang2024telling}} &
      \textbf{\methodname~(Ours)} \\
    \end{tabular}
   } \\
   \includegraphics[valign=c, width=\linewidth]{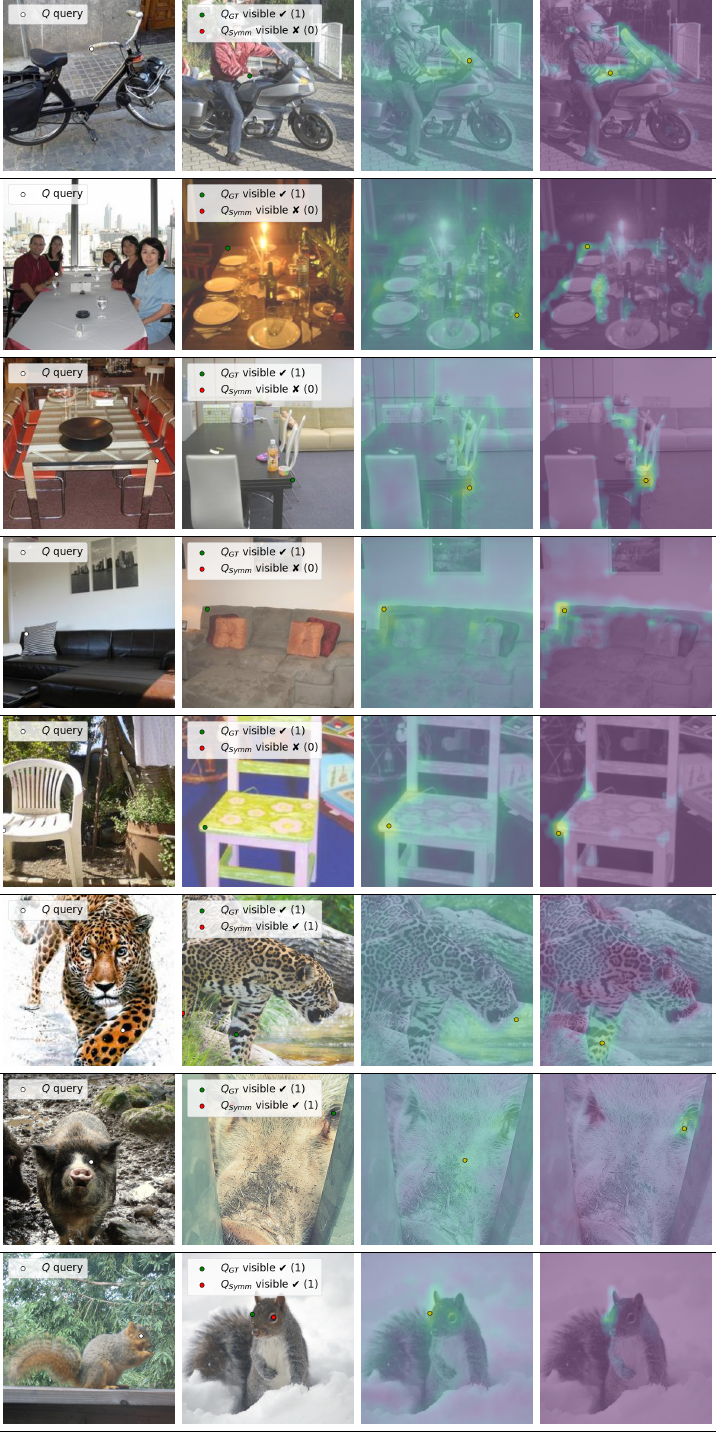} 
  \caption{\textbf{Qualitative results on the task of assignment to the correct correspondence (11-case) on PFPascal~\cite{ham2017proposal} (First five rows), APK~\cite{zhang2024telling} (In the middle) and CUB~\cite{wah2011caltech} (Last row).} In the samples above the symmeric counterpart is visible in the target view. While previous work assigns attention on most of the image, our model is more confident in the predictions and only attends to the regions of interest.}\label{fig:pck_11_failure}
\end{figure}

\begin{figure}[ht!]
    \centering
    \footnotesize
    \setlength{\tabcolsep}{0pt}
    \makebox[\linewidth][c]{%
      \begin{tabular}{*{4}{C{0.25\linewidth}}}
        \textbf{Source $S$} &
        \textbf{Target $T$} &
        \textbf{Geo~\cite{zhang2024telling}} &
        \textbf{\methodname~(Ours)} \\
      \end{tabular}
     } \\
     \includegraphics[valign=c, width=\linewidth]{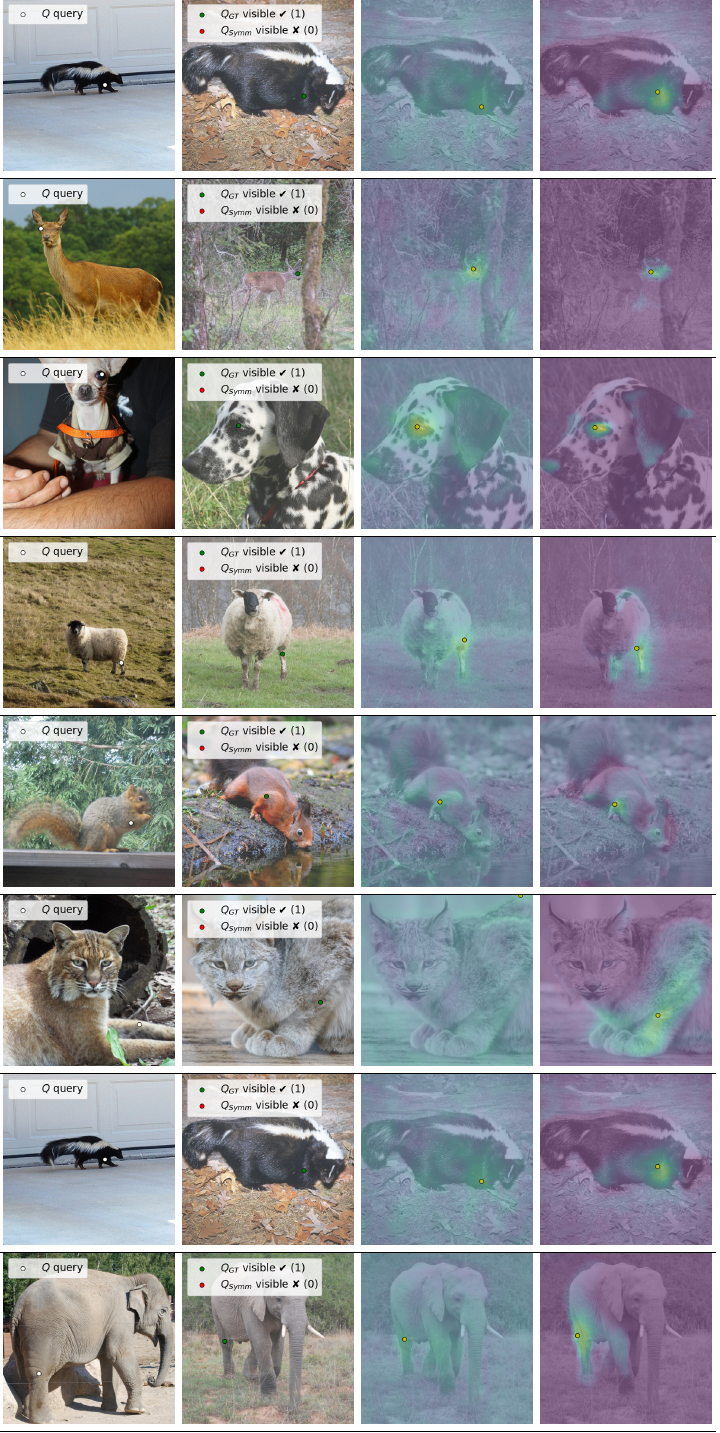} 
    \caption{\textbf{Qualitative results on the task of semantic correspondence estimation (10-case) on APK~\cite{zhang2024telling} and CUB~\cite{wah2011caltech} (Last row).} In the above samples only one semantically similar keypoint is visible in the target view.
    While previous work assigns attention on most of the image, our model is more confident in the predictions and only attends to the regions of interest.}\label{fig:pck_10_failure}
  \end{figure}

We also report performance on the geometrically relevant case, where multiple semantically similar keypoints are present in the target image. In this scenario, our method is more confident in the predictions, focusing only on the relevant areas of the image (see \cref{fig:pck_11_failure}). 
As shown in \cref{fig:pck_10_failure}, our features localize keypoints precisely in the target image without spreading similarity across irrelevant regions.

\paragraph{PGCK Dataset imbalance}
We analyze the cardinality of the geometrically aware subset in \cref{fig:subsets_PCK}, which is able to assess the confusion between keypoints on opposite sides of the symmetry axis. To complete the table for all categories, we present a detailed report for the CUB~\cite{wah2011caltech}, SPair~\cite{min2019spair}, and APK~\cite{zhang2024telling} datasets in \cref{fig:subsets_PCK_detail}.


\begin{table}[t!]
  \footnotesize
  \centering 
  \begin{tabularx}{1\linewidth}{*{9}{X}}
    \hline
    \tabularxmulticolumncentered{9}{c}{\textbf{CUB}~\cite{wah2011caltech}}\\
    \hline
    &&&&Bir\\
    &&&&11\\
    \hline
    \tabularxmulticolumncentered{9}{c}{\textbf{SPair}~\cite{min2019spair}}\\
    \hline
    Aer&Bik&Bir&Boa&Bot&Bus&Car&Cat&Cha\\
    34&53&23&30&97&43&40&82&81\\
    Cow&Dog&Hor&Mot&Per&Pla&She&Tra&TV\\
    80&46&61&69&34&73&72&90&62\\
    \hline
    \tabularxmulticolumncentered{9}{c}{\textbf{APK}~\cite{min2019spair} }\\
    \hline
  alo&ant&arg&bea&bis&bla&bob&bro&buf\\
  37&31&28&21&20&32&36&33&34\\
  cat&che&chi&cow&dee&dog&ele&fox&gir\\
  42&30&37&38&36&36&36&33&24\\
  gor&ham&hip&hor&jag&kin&leo&lio&mar\\
  42&34&40&26&32&33&40&40&27\\
  mon&moo&mou&noi&ott&pan&pan&pig&pol\\
  37&24&20&38&28&23&46&35&30\\
  rab&rac&rat&rhi&she&sku&sno&spi&squ\\
  7&28&27&23&34&19&35&35&16\\
  tig&uak&wea&wol&zeb\\
  32&39&36&34&24\\    
  \end{tabularx}
  \caption{\textbf{PGCK Dataset imbalance}. Completing \cref{fig:subsets_PCK}, we report the geometric aware $\nicefrac{n_{11}}{n}$, counts only the pairs for which a geometric mismatch is possible. Categories often present high imbalance, and geometrical error modes would have a different weight.}\label{fig:subsets_PCK_detail}
\end{table}

\clearpage
\subsection{Feature Space Segmentation}\label{secsup:experiments:05_experiments_02_segmentation}
In \cref{sec:experiments:segmentation}, we revisit the procedure for evaluating the separability of the feature space by leveraging annotated parts and computing centroids for each. To complement the qualitative analysis, we now present a quantitative evaluation using standard segmentation metrics. Furthermore, we include normalized confusion matrices for another qualitative analysis to highlight the geometric consistency of the predicted segmentations.

\paragraph{Dataset}
We use the PascalParts~\cite{chen2014detect} dataset, which offers detailed part annotations for a wide variety of object categories, enabling consistent and comprehensive evaluation across different classes. However, some of the 20 categories, such as \textit{boat, table, and sofa}, have only a single part annotated. As a result, we exclude these categories from our evaluation.

\paragraph{Evaluation}
Similar to prior work~\cite{peyre2019computational}[\hyperlink{bib:oliveira2016deep}{68}] we use the \textbf{accuracy} and \textbf{mean Intersection over Union (mIoU)} as our primary evaluation metrics. To compute these metrics, let \( s_{ij} \) represent the number of patches from ground-truth part \( i \) that are predicted as part \( j \) by the model.

The mean Intersection over Union (mIoU) is defined as
\begin{equation}
    \text{mIoU} = \frac{1}{N} \sum_i \frac{s_{ii}}{\sum_j s_{ij} + \sum_j s_{ji} - s_{ii}},
\end{equation}
where $N$ is the number of parts.
The accuracy is defined as:
\begin{equation}
    \text{Acc} =   \frac{\sum_i s_{ii}}{\sum_i \sum_j s_{ij}}.
\end{equation}

This approach provides a per-category measure, enabling a detailed comparison of part segmentation performance across different parts. For an overall evaluation, we compute the average metric across all categories.

In addition, we report metrics specifically for the geometric subdivision, which includes parts divided into left and right counterparts, such as \textit{legs, wings, eyes}. The geometric subdivision is presented both quantitatively and qualitatively through a confusion matrix. This focused analysis allows us to assess the model’s ability to distinguish symmetric parts, which is critical for understanding its geometric reasoning capabilities.

\paragraph{Results}
In \cref{tab:clustering}, we report the average mIoU and average accuracy across categories, evaluating performance on the segmentation task using the PascalParts~\cite{chen2014detect} dataset.

\begin{table}[t!]
    \footnotesize
    \centering 
    \begin{tabularx}{1\linewidth}{m{1.7cm} *{4}{Y}}
        & {mean mIoU}$\uparrow$ & {mean Acc}$\uparrow$ & {mean mIoU$\uparrow$ (geo)} & {mean Acc$\uparrow$ (geo)}\\
          \hline
          {DINO~\cite{caron2021emerging}} & 31.5 & 87.9 & 44.1 & 95.9 \\
          {DIFT$_{\textit{adm}}$~\cite{tang2023emergent}} & 31.5 & 87.9 & 48.0 & 96.4 \\
          {DINOv2B~\cite{oquab2023dinov2}} & \textbf{39.8} & \textbf{90.1} & 57.6 & 97.3 \\
          \hline
          {Geo~\cite{zhang2024telling}} & 37.1 & 88.4 & \textbf{62.8} & \textbf{97.6} \\
          {\methodname} {(Ours)} & \underline{37.9} & \underline{89.0} & \textbf{63.0} &  \textbf{97.5} \\
    \end{tabularx}
    \caption{\textbf{Quantitative Evaluation of Segmentation Confusion Matrices on PascalParts~\cite{chen2014detect}.} Our features retain the segmentation performance of DINOv2 (two left columns), while additionally enabling the distinction between left and right parts (right two columns). The best scores are shown in \textbf{bold} and the second best are \underline{underlined}. Methods with a performance difference of less than 0.5\% are considered to be on par.}\label{tab:clustering}
  \end{table}
  First, evaluating all parts within each category on the left side of \cref{tab:clustering}, offers insight into the model’s capacity to differentiate between parts that are not necessarily symmetric, such as  \textit{head, body, and tail}.
  Although accuracy remains high across methods, mIoU scores decline due to sensitivity to class imbalance in the parts. Both fine-tuned geometry aware methods (last two rows) perform worse on non geometric parts, as expected. However, our method outperforms Geo~\cite{zhang2024telling}, indicating better retention of non geometric information.

\begin{figure}[t!]
    \centering
    \vspace{1em}
    \begin{overpic}[angle=0,trim= 0.0cm 0.0cm 0cm 0.0cm,clip, width=0.88\linewidth]{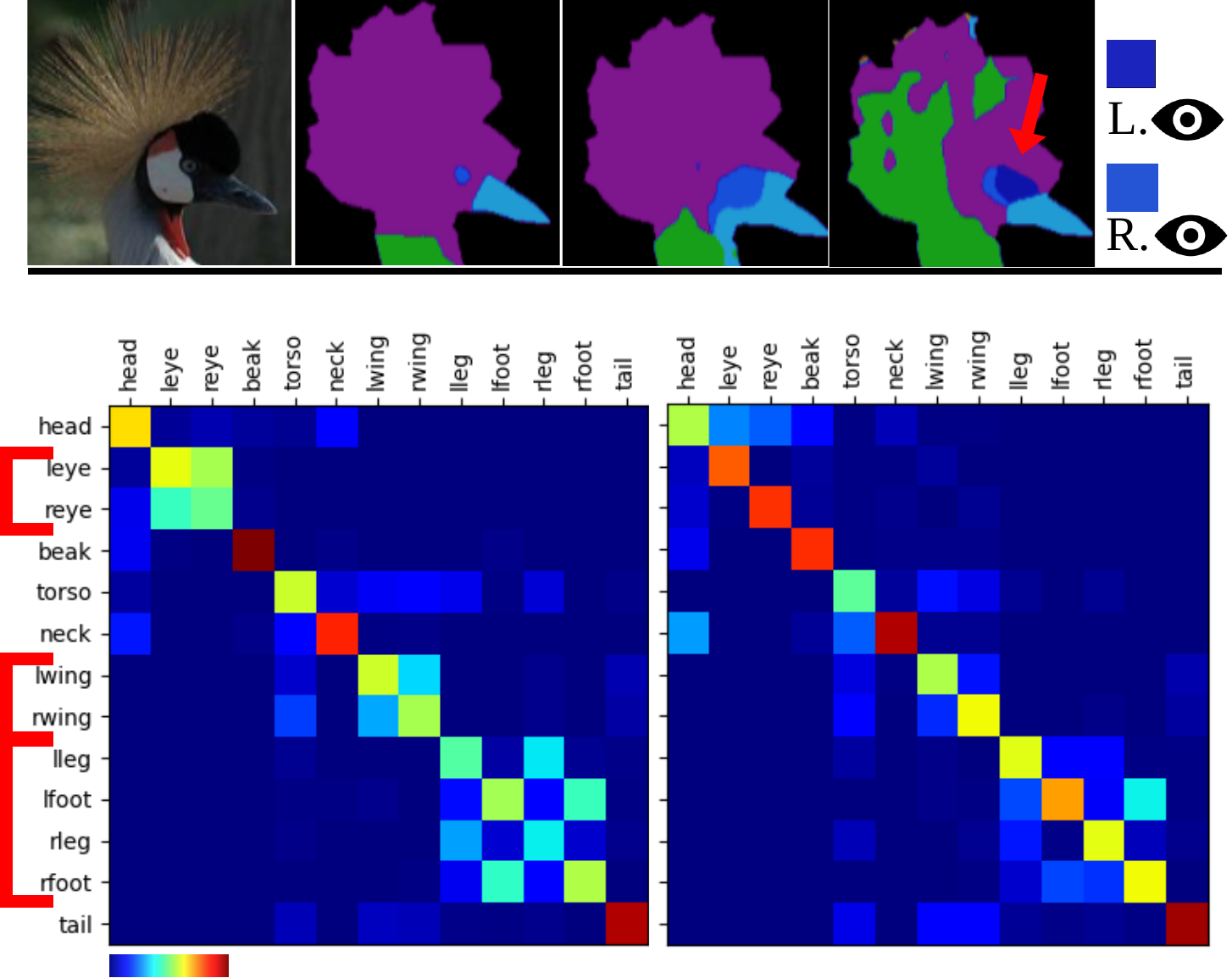}
        \scriptsize
        \put(22,53.5){DINOv2-B~\cite{oquab2023dinov2}}
        \put(66,53.5){GECO (Ours)}
    
        \put(10,81){Input}
        \put(32,81){GT}
        \put(47,81){GECO (Ours)}
        \put(68,81){DINOv2-B~\cite{oquab2023dinov2}}
    
        \put(40,-0.5){Part Confusion Matrices}
    
        \put(7,0){\tiny 0}
        \put(20,0){\tiny Max}
    \end{overpic}
    \caption{\textbf{Results on Symmetry.}  \textbf{Top:} Our method reduces symmetric mismatches in part clustering. \textbf{Bottom:} We show the part confusion matrix (patches aggregated for each part across images), highlighting geometric confusion (red brackets).}
    \label{fig:segmentation_conf_quali_relation}

    \end{figure}

  On the right side of \cref{tab:clustering}, we report the metrics for the geometric parts only. As expected, both geometry-aware methods outperform all foundation models.
 The connection between the qualitative part segmentation results in \cref{fig:clustering1} and the confusion matrix analysis is demonstrated in \cref{fig:segmentation_conf_quali_relation}.
  Visual inspection aligns with the numbers and reveals that DINOv2-B struggles with symmetric distinctions (e.g., left vs. right).
  Additional examples of checkerboard patterns indicative of geometric confusion are shown in \cref{fig:clustering_confusion}. Here, the matrices contain only geometrically confusable parts and are normalized to mitigate class imbalance.

  \clearpage

\newcommand{\QualitativeRowSEG}[2]{
    \hspace{0.5cm}
    \ifthenelse{\equal{#1}{aeroplane}}{
      \incimicolor{figures_sup/05_experiments/seg/pascalparts/#1/dinov2_b14_518_upft1/conf/test/geo/colourbar_conf_matrix_normalized.png}
    }{
    }
    &
    \vspace{-1.5cm} #2 
    &
    \incimilabel{figures_sup/05_experiments/seg/pascalparts/#1/dinov2_b14_518_upft1/conf/test/geo/matrix_conf_matrix_normalized.png}
    &\incimi{figures_sup/05_experiments/seg/pascalparts/#1/dino_224_upft4/conf/test/geo/matrix_conf_matrix_normalized.png}
    &\incimi{figures_sup/05_experiments/seg/pascalparts/#1/dift_adm/conf/test/geo/matrix_conf_matrix_normalized.png}
    &\incimi{figures_sup/05_experiments/seg/pascalparts/#1/dinov2_b14_518_upft1/conf/test/geo/matrix_conf_matrix_normalized.png}
    &\incimi{figures_sup/05_experiments/seg/pascalparts/#1/\Geo/conf/test/geo/matrix_conf_matrix_normalized.png}
    &\incimi{figures_sup/05_experiments/seg/pascalparts/#1/geco/conf/test/geo/matrix_conf_matrix_normalized.png}\\
}
\begin{figure*}[ht!]
    \footnotesize
    \newcolumntype{A}{>{\raggedright\arraybackslash\hsize=0.3\hsize} Y}
    \newcolumntype{Z}{>{\raggedleft\arraybackslash\hsize=0.3\hsize} Y}
    \newcolumntype{B}{>{\raggedleft\arraybackslash\hsize=0.3\hsize} Y}
    \begin{tabularx}{\linewidth}{A Z B *{5}{Y}}
      &&&\textbf{DINO~\cite{caron2021emerging}}
      &\textbf{DIFT$_{\textit{adm}}$~\cite{tang2023emergent}}
      &\textbf{DINOv2-B~\cite{oquab2023dinov2}}
      &\textbf{Geo~\cite{zhang2024telling}}
      &\textbf{\methodname} \textbf{(Ours)} \\
      \QualitativeRowSEG{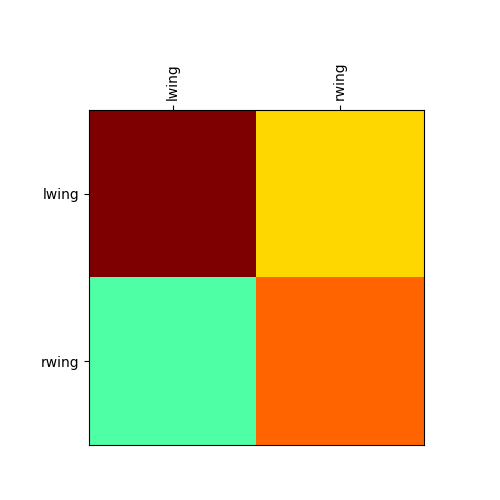}{aero}
      \QualitativeRowSEG{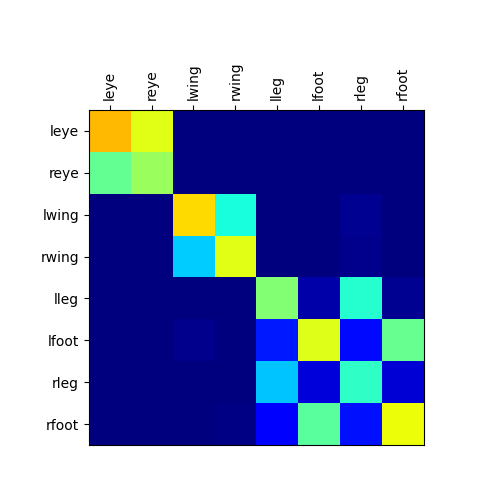}{bird}
      \QualitativeRowSEG{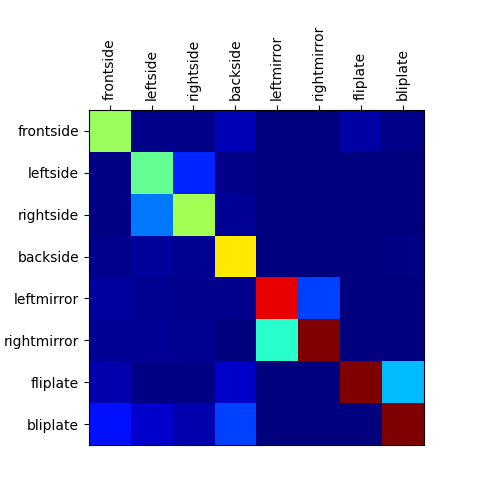}{bus}
      \QualitativeRowSEG{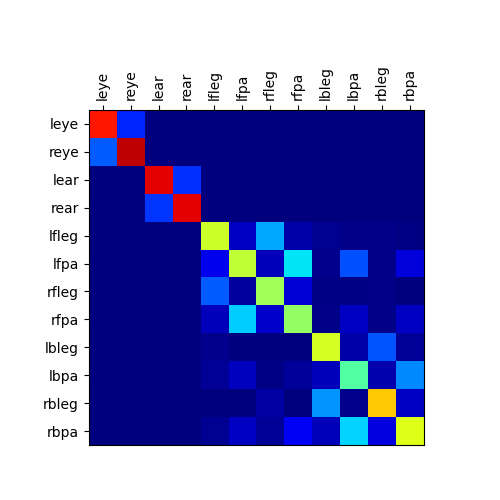}{cat}
      \QualitativeRowSEG{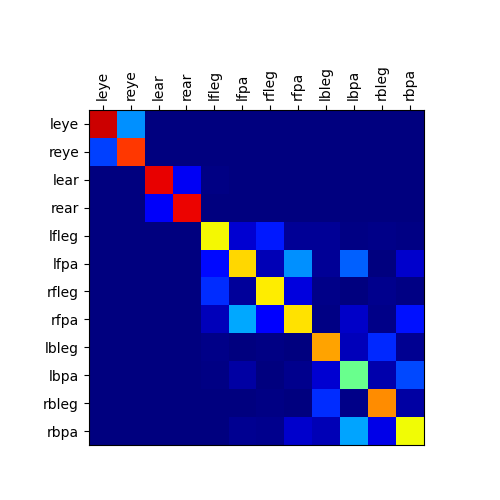}{dog}
      \QualitativeRowSEG{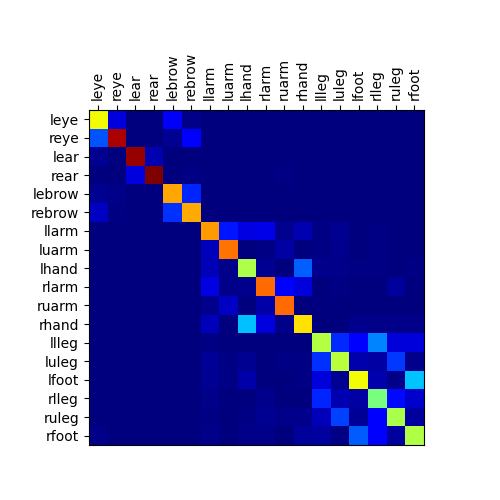}{person}
    \end{tabularx}
    \caption{\textbf{Qualitative Evaluation of Segmentation Confusion Matrices on PascalParts~\cite{chen2014detect}.} We present the confusion matrix for the segmentation task on the PascalParts~\cite{chen2014detect} dataset, showing every second category in alphabetical order. The matrix displays predicted parts on vertical and ground truth parts on the horizontal axis, plotting only parts per category with annotated symmetrical counterparts for visualization purposes. Diagonal entries indicate correct predictions, while off-diagonal entries show misclassifications. To address class imbalance, the matrix is column-normalized: each column is divided by the total number of patches for that ground truth part. With number of parts $N$, each ground truth part annotation is assigned a total mass of $\nicefrac{1}{N}$, which is distributed across the predicted parts, such that the columns sum to $\nicefrac{1}{N}$. The colour coding depends on the number of parts $N$.
    }\label{fig:clustering_confusion}
    \vspace{3em}
\end{figure*}

\subsection{2D-3D Matching}\label{secsup:experiments:05_experiments_03_pnp}

Structure-from-Motion (SfM), Perspective-n-Point (PnP), and related methods that rely on rigid geometry are not typical applications of non-rigid intra-category feature learning, which remains effective even under substantial intra-class variations in shape and appearance. Nevertheless, we demonstrate that our features support downstream tasks such as inter-instance 2D-3D alignment, which is essential for pose estimation and geometric reasoning.

\begin{figure}[ht!]
  \centering
  \begin{overpic}[angle=0,trim= 0.0cm 0.0cm 0cm 0.0cm,clip, width=1\linewidth]{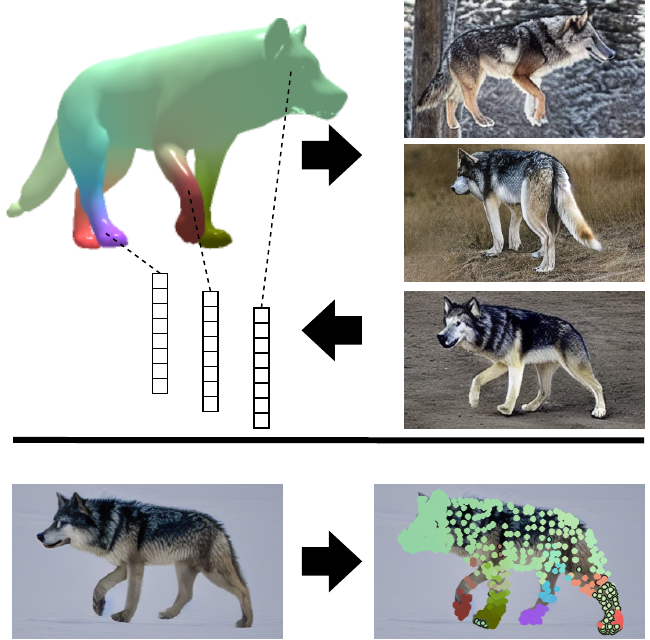}
      \scriptsize
      \put(2.5,95){\large{\underline{\textbf{Populate Mesh}}}}
      \put(2.5,26){\large{\underline{\textbf{Test}}}}
      \put(50,68){\makebox(0,0){Depth-}}
      \put(50,64){\makebox(0,0){Guided}}
      \put(50,60){\makebox(0,0){ControlNet}}      
      \put(50,42){\makebox(0,0){Median}}
      \put(50,38){\makebox(0,0){Vertex}}
      \put(50,34){\makebox(0,0){Features}}
      \put(60,26){geodesic and rotation error (PnP)}
  \end{overpic}
  \caption{\textbf{Evaluation Protocol.}
  \textbf{Top:} A mesh from the SMAL model~[\protect\hyperlink{bib:zuffi3DMenagerieModeling2017}{69}] is rendered to obtain depth images, which guide ControlNet-based synthesis of realistic 2D images. Hallucinated outputs are manually filtered. Mesh vertices are then populated with features by taking the median across the corresponding 2D views. 
  \textbf{Bottom:} A novel view is processed, and 2D-3D correspondences are established via argmax matching to the mesh vertex with the highest feature similarity. Matches are color-coded according to surface color. Those exceeding a geodesic distance threshold ${d}_\mathrm{geo}$ to the ground-truth location are highlighted with a black contour.
  }\label{fig:2D3D:geodesic_dist}
\end{figure}

\paragraph{Dataset}
To obtain data that combines intra-class variability with rigid structure, we render two 3D animal meshes from the SMAL model~[\hyperlink{bib:zuffi3DMenagerieModeling2017}{69}]. We enhance their realism and intra-class diversity using ControlNet (see \cref{fig:2D3D:geodesic_dist}, Top), and j filter out hallucinated outputs.

\paragraph{Procedure}
We decorate all mesh vertices with our features by median-aggregating across views, using the ground-truth camera poses (see \cref{fig:2D3D:geodesic_dist}, Top). For 20 novel images, we compute 2D-3D correspondences (see \cref{fig:2D3D:geodesic_dist}, Bottom) using argmax matching to the mesh vertex with the highest feature similarity.

We report the mean geodesic distance between matched 2D keypoints (unprojected onto the mesh) $p$ and ground-truth vertex $p_{\text{gt}}$:
\begin{equation}
\bar{d}_\mathrm{geo} = \mathrm{mean}\left(d_\mathrm{geo}(p, p_{\text{gt}})\right),
\end{equation}
where $d_\mathrm{geo}$ denotes the shortest path along the mesh surface between two vertices. In \cref{fig:2D3D:geodesic_dist} (Bottom), black contours indicate regions of high geodesic error, where the matched 3D vertex is distant from the ground truth.

We further evaluate camera pose estimation based on these 2D-3D matches. A robust PnP algorithm inside RANSAC estimates the camera's rotation and translation relative to the 3D object. The quality of the estimated rotation $R$ is assessed by comparing it to the ground-truth rotation $R_{\text{gt}}$ using the median rotation error across all images:
\begin{equation}
\bar{e}_\mathrm{rot} = \mathrm{median}\left(\arccos\left(\frac{\mathrm{trace}(R{\text{gt}}^\top R) - 1}{2}\right)\right).
\end{equation}

\paragraph{Results}
As shown in \cref{tab:2D3D}, our \methodname features outperform both Geo~\cite{zhang2024telling} and DINOv2-B~\cite{oquab2023dinov2} in terms of rotation error and geodesic distance. This confirms that our method enables effective intra-category 2D-3D matching for pose estimation and geometric reasoning.
\begin{table}[ht!]
    \footnotesize
    \centering 
    \begin{tabularx}{1\linewidth}{m{1.9cm} *{2}{Y}}
          &$\bar{e}_\mathrm{rot} (^\circ) \downarrow$  & $\bar{d}_\mathrm{geo} \downarrow$\\
          \hline
          DINOv2-B~\cite{oquab2023dinov2}  & 14.8 & 0.34\\
          \hline
          Geo~\cite{zhang2024telling} & 10.4& 0.35 \\
          {\methodname (Ours)}  &\textbf{7.0}& \textbf{0.24} \\
        \end{tabularx}
        \caption{\textbf{Quantitative Evaluation of Viewpoint Reconstruction and 2D-3D Matching.} 
        We assess the quality of 2D-3D correspondences using the median rotation error $\bar{e}_\mathrm{rot}$ of the reconstructed view and mean geodesic distance $\bar{d}_\mathrm{geo}$ between matched and ground-truth keypoints. Our method outperforms DINOv2-B~\cite{oquab2023dinov2} and Geo~\cite{zhang2024telling} on both metrics, demonstrating superior performance in pose estimation and geometric reasoning. Best scores are shown in \textbf{bold}.
        }\label{tab:2D3D}
   \vspace{3cm}
\end{table}
\clearpage
{
    \small

      \makeatletter
      \renewcommand\@biblabel[1]{[\number\numexpr#1+66\relax]}
      \makeatother

}
\clearpage
\clearpage

\begin{figure*}
    \centering
    \footnotesize
    \setlength{\tabcolsep}{0pt}
    \makebox[\linewidth][c]{%
      \begin{tabular}{*{7}{C{0.142857143\linewidth}}}
        \textbf{Source $S$}
        &\textbf{Target $T$}
        &\textbf{DINO~\cite{caron2021emerging}} 
        &\textbf{DIFT$_{\text{ad}}$~\cite{tang2023emergent}} 
        &\textbf{DINOv2-B~\cite{oquab2023dinov2}} 
        &\textbf{Geo \cite{zhang2024telling}}
        & \textbf{\methodname} \textbf{(Ours)} \\
      \end{tabular}
     }\\
     \includegraphics[valign=c, width=\linewidth]{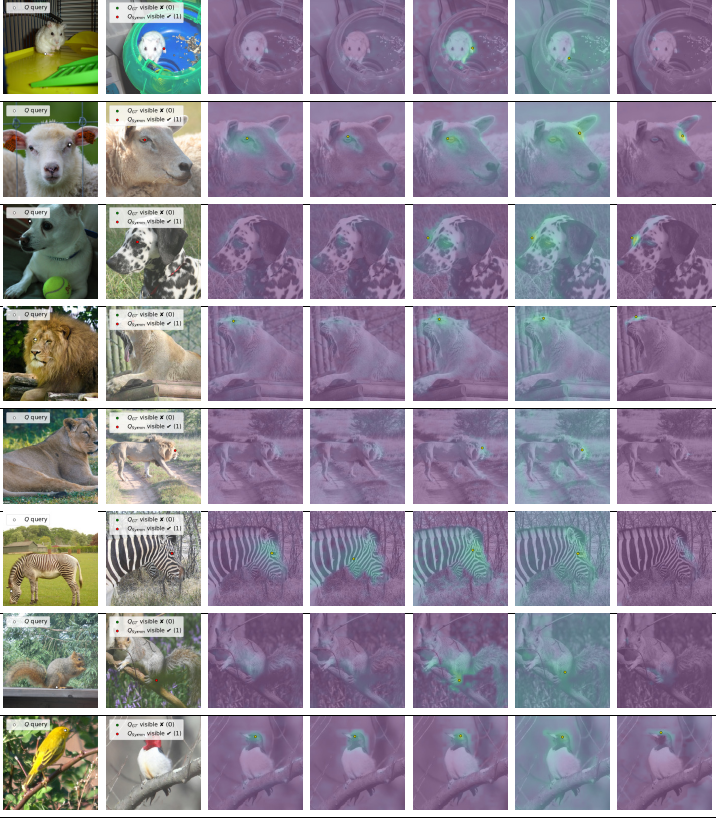} 
    \caption{\textbf{Qualitative results on the task of assignment to the bin (01-case) on APK~\cite{zhang2024telling} and CUB~\cite{wah2011caltech} (Last row).}}\label{fig:pck_quali01}
\end{figure*}


\begin{figure*}
    \centering
    \footnotesize
    \setlength{\tabcolsep}{0pt}
    \makebox[\linewidth][c]{%
      \begin{tabular}{*{7}{C{0.142857143\linewidth}}}
        \textbf{Source $S$}
        &\textbf{Target $T$}
        &\textbf{DINO~\cite{caron2021emerging}} 
        &\textbf{DIFT$_{\text{ad}}$~\cite{tang2023emergent}} 
        &\textbf{DINOv2-B~\cite{oquab2023dinov2}} 
        &\textbf{Geo \cite{zhang2024telling}}
        & \textbf{\methodname} \textbf{(Ours)} \\
      \end{tabular}
     }\\
     \includegraphics[valign=c, width=\linewidth]{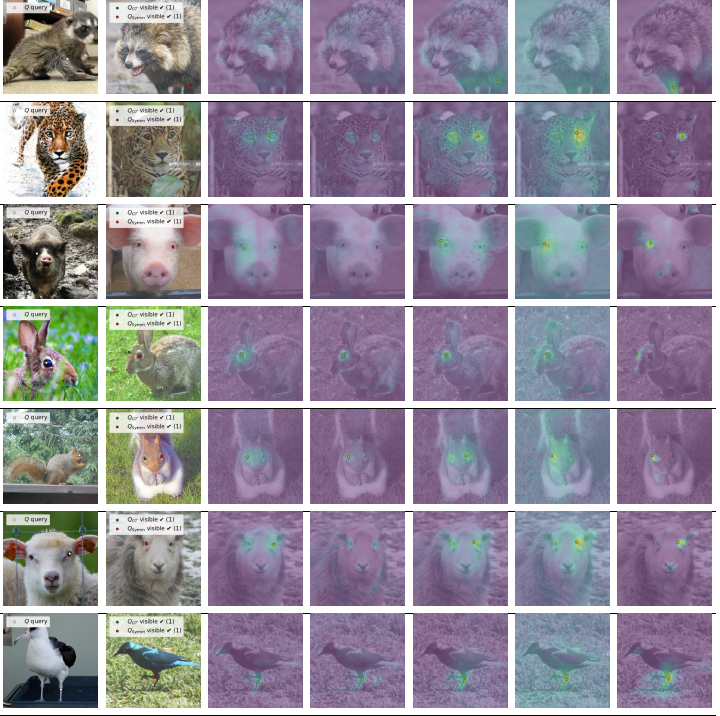} 
     \caption{\textbf{Qualitative results Qualitative results on the task of assignment to the correct correspondence (11-case) on APK~\cite{zhang2024telling}, and CUB~\cite{wah2011caltech} (Last row). Note that the keypoint pair in the last row would be excluded in the unambiguous TP subdivision.}}\label{fig:pck_quali11}
\end{figure*}


\begin{figure*}
    \centering
    \footnotesize
    \setlength{\tabcolsep}{0pt}
    \makebox[\linewidth][c]{%
      \begin{tabular}{*{7}{C{0.142857143\linewidth}}}
        \textbf{Source $S$}
        &\textbf{Target $T$}
        &\textbf{DINO~\cite{caron2021emerging}} 
        &\textbf{DIFT$_{\text{ad}}$~\cite{tang2023emergent}} 
        &\textbf{DINOv2-B~\cite{oquab2023dinov2}} 
        &\textbf{Geo \cite{zhang2024telling}}
        & \textbf{\methodname} \textbf{(Ours)} \\
      \end{tabular}
     }\\
     \includegraphics[valign=c, width=\linewidth]{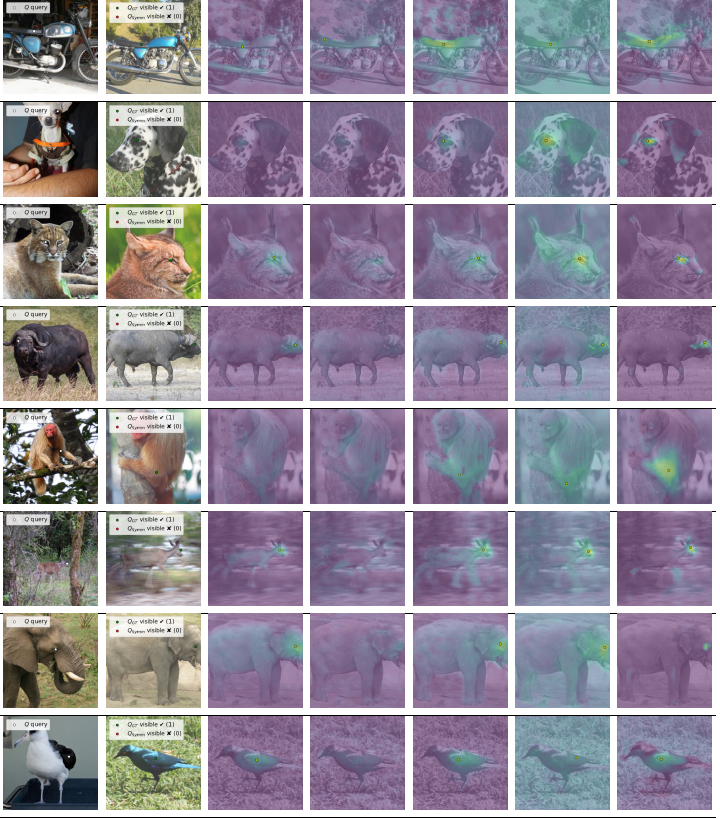} 
     \caption{\textbf{Qualitative results Qualitative results on the task of semantic correspondence estimation (10-case) on PFPascal~\cite{ham2017proposal} (First row), APK~\cite{zhang2024telling}, and CUB~\cite{wah2011caltech} (Last row).}}\label{fig:pck_quali10}
\end{figure*}

\end{document}